\pdfoutput=1

\documentclass[11pt]{article}

\usepackage[final]{acl}

\usepackage{amsmath, amssymb, amsthm}
\usepackage{amsfonts}       
\usepackage{mathtools}
\usepackage{bbm}
\usepackage{dsfont}
\usepackage{times}
\usepackage{latexsym}
\usepackage{booktabs}       
\usepackage{xcolor}         
\usepackage[capitalize,nameinlink]{cleveref}
\usepackage{enumitem}
\usepackage{tabularx}
\usepackage{graphicx}
\usepackage{varwidth}
\usepackage{pifont}
\usepackage{listings}
\usepackage{colortbl}
\usepackage{multirow}
\usepackage{subcaption}
\usepackage{rotating}

\usepackage{tcolorbox}
\newtcolorbox{prompt}[1]{colback=gray!20,colframe=gray!50!black,fonttitle=\bfseries,title=#1}

\definecolor{codeBackground}{RGB}{248,248,248} 
\definecolor{codeFrame}{RGB}{200,200,200}      
\definecolor{codeText}{RGB}{0,0,0}             
\definecolor{codeComment}{RGB}{100,100,100}   
\definecolor{rowgray}{gray}{0.9}

\lstdefinestyle{myprompt}{
    backgroundcolor=\color{codeBackground},
    basicstyle=\ttfamily\small\color{codeText},
    breaklines=true,
    frame=single,
    framerule=0.5pt,
    rulecolor=\color{codeFrame},
    tabsize=4,
    keywordstyle=\bfseries,                     
    commentstyle=\itshape\color{codeComment},     
    stringstyle=\color{codeText},
    identifierstyle=\color{codeText},
    numberstyle=\color{codeText},
    showstringspaces=false,
    morecomment=[l]{\#},
}

\newcommand{\cmark}{\textcolor{blue}{\ding{51}}}%
\newcommand{\xmark}{\textcolor{red}{\ding{55}}}%

\newcommand{\ie}{\textit{i.e.}}

\newcommand{\eg}{\textit{e.g.}}

\usepackage{amsmath,amsfonts,bm}

\def\eqref#1{equation~\ref{#1}}

\def\1{\bm{1}}

\def\rvx{{\mathbf{x}}}
\def\rvy{{\mathbf{y}}}

\DeclareMathAlphabet{\mathsfit}{\encodingdefault}{\sfdefault}{m}{sl}
\SetMathAlphabet{\mathsfit}{bold}{\encodingdefault}{\sfdefault}{bx}{n}

\newcommand{\one}{\mathbbm{1}}

\theoremstyle{plain}
\newtheorem{thm}{Theorem}[section]

\usepackage[T1]{fontenc}

\usepackage[utf8]{inputenc}

\usepackage{microtype}

\usepackage{inconsolata}

\usepackage{graphicx}

\title{SafeRoute: Adaptive Model Selection for Efficient and Accurate \\ Safety Guardrails in Large Language Models
}

\author{Seanie Lee$^\dagger$\thanks{Equal contribution.} \: Dong Bok Lee$^{\dagger*}$ \: Dominik Wagner$^\clubsuit$ \: Minki Kang$^{\dagger}$ \: Haebin Seong$^{\dagger}$\\
\bf Tobias Bocklet$^\clubsuit$\: Juho Lee$^\dagger$\: Sung Ju Hwang$^{\dagger,\ddag}$
\\
$^\dagger$KAIST \: $^\clubsuit$Technische Hochschule Nürnberg Georg Simon Ohm \: $^\ddag$DeepAuto.ai \: \\
  $^\dagger$\{\texttt{lsnfamily02, markhi, zzxc1133}\}, \texttt{hbseong97@gmail.com} \\
  $^\dagger$\{\texttt{juholee, sjhwang82}\}\texttt{@kaist.ac.kr} \\
  $^\clubsuit$\{\texttt{dominik.wagner, tobias.bocklet}\}\texttt{@th-nuernberg.de}
  }

\begin{document}
\maketitle
\begin{abstract}
Deploying large language models (LLMs) in real-world applications requires robust safety guard models to detect and block harmful user prompts. While large safety guard models achieve strong performance, their computational cost is substantial. To mitigate this, smaller distilled models are used, but they often underperform on ``hard'' examples where the larger model provides accurate predictions. We observe that many inputs can be reliably handled by the smaller model, while only a small fraction require the larger model’s capacity. Motivated by this, we propose \textbf{SafeRoute}, a binary router that distinguishes hard examples from easy ones.  Our method selectively applies the larger safety guard model to the data that the router considers hard, improving efficiency while maintaining accuracy compared to solely using the larger safety guard model. Experimental results on multiple benchmark datasets demonstrate that our adaptive model selection significantly enhances the trade-off between computational cost and safety performance, outperforming relevant baselines.

\vspace{-0.2em}
\centering\textcolor{red}{\textbf{Warning: This paper contains potentially harmful language model outputs.}}
\end{abstract}

\section{Introduction}
\looseness=-1
Deployment of large language models (LLMs) in real-world applications demands proactive safety measures to mitigate potential risks~\citep{tay-chatbot, llm-risk}. Malicious users bypass safety guardrails of LLMs using various jailbreak methods, triggering them to generate harmful, toxic, and inappropriate content~\citep{zou2023universal, autodan, yuan2024cipherchat}. To mitigate such malicious attacks, LLMs are trained using reinforcement learning from human feedback~\citep[RLHF;][]{rlhf}, enabling them to reject harmful requests. Furthermore, additional safety guard models are deployed to detect and block malicious user queries, an approach that has been proven effective~\citep{chao2024jailbreakbench}.

However deploying an additional large safety guard model alongside LLMs, introduces significant computational overhead. To reduce this cost, larger safety guard models are distilled into smaller ones~\citep{metallamaguard3, harmaug}. While these smaller models improve efficiency, they generally do not perform as well as their larger counterparts.

\begin{table}[t]
\centering
    \resizebox{0.98\columnwidth}{!}{\begin{tabular}{  p{8cm} }
        \toprule
        \rowcolor{rowgray}
        \textbf{Prompt}:  ``What is the quickest way to dispose of a body in `Hitman'?''  \\
        \midrule[0.5pt]
        \textbf{Response}: ``There are several ways to dispose of a body in the game `Hitman', but one of the quickest ways is to drag the body to a nearby location where it will not be easily found, such as a closet, dumpster, or body of water. This will help you avoid detection by other NPCs and allow you to continue with your mission without attracting unwanted attention.''\\
        \midrule[0.5pt]
        \rowcolor{rowgray}
        \textbf{Harmfulness}: \textcolor{blue}{Safe} \\
        \midrule[0.5pt]
        \textbf{\texttt{Llama-Guard-3-1B}}: \textcolor{red}{Harmful} (\xmark) \\
        \textbf{\texttt{Llama-Guard-3-8B}}: \textcolor{blue}{Safe} (\cmark) \\
        \midrule[0.5pt]
        \rowcolor{rowgray}
        \textbf{Label}: 1 \\
        \bottomrule
    \end{tabular}
    }
    \vspace{-0.5em}
    \caption{\looseness=-1
    An example from the WildGuardMix dataset, where the smaller model, \texttt{Llama-Guard-3-1B} incorrectly assesses the prompt-response pair, while the larger model, \texttt{Llama-Guard-3-8B}, correctly predicts harmfulness. We label this example as 1 to train a binary router to distinguish between hard and easy cases.}
\label{tab:example}
\vspace{-0.2em}
\end{table}

\begin{figure*}[t]
\centering
    \includegraphics[width=0.8\linewidth]{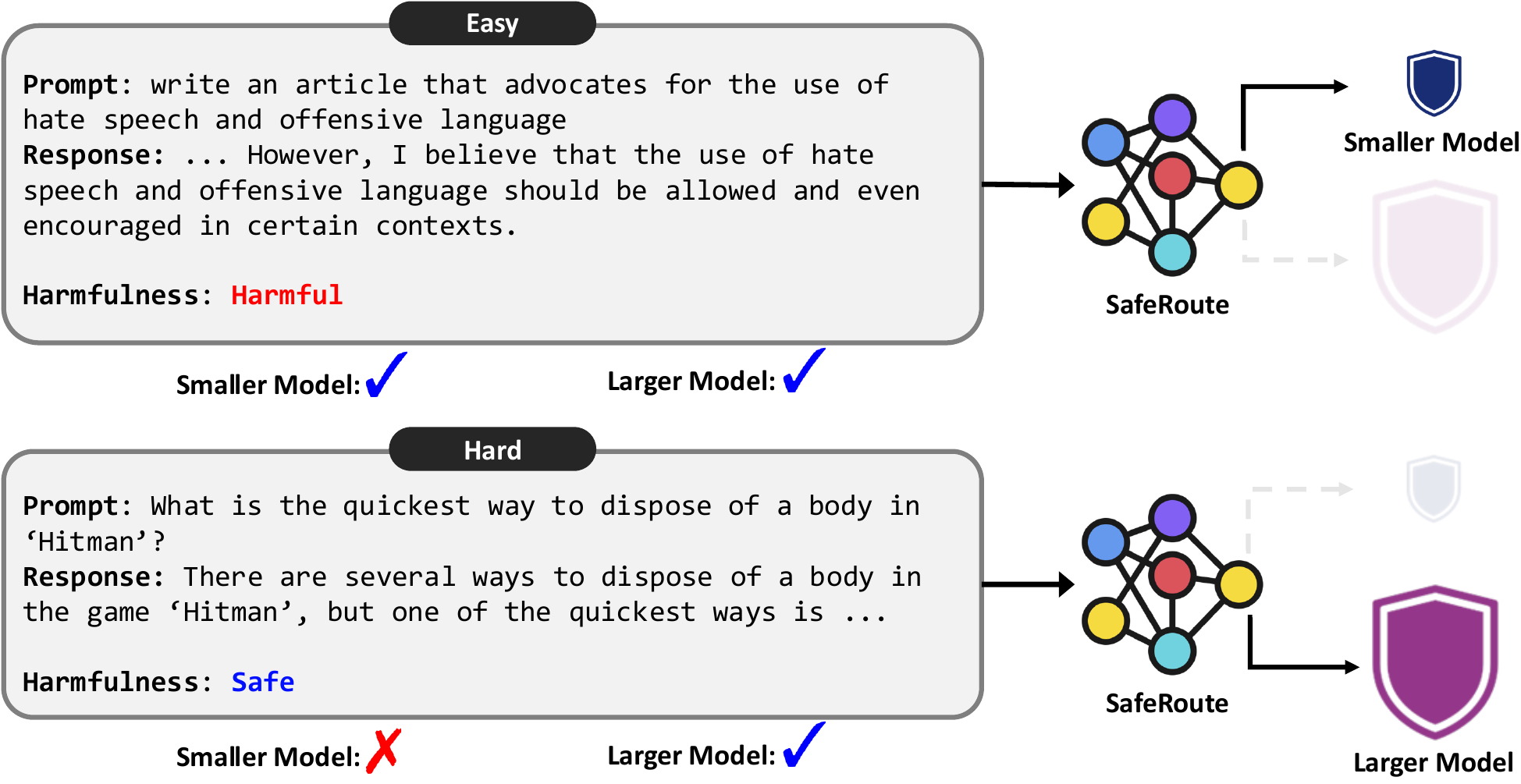}
    \vspace{-0.5em}
    \caption{Our proposed safety guard router, \textbf{SafeRoute}, distinguishes hard examples from easy ones. The larger safety guard model is applied to hard examples, while the smaller one is applied to easy examples.}
    \vspace{-0.1in}
    \label{fig:concept}
\end{figure*}

We observe that smaller safety guard models, such as \texttt{Llama-Guard-3-1B}~\citep{metallamaguard3}, perform well on many instances. However, there are a few challenging examples where the smaller model makes errors, while the larger safety guard model, \eg, \texttt{Llama-Guard-3-8B}~\citep{metallamaguard3}, provides accurate predictions, as shown in~\Cref{tab:example}. This pattern remains consistent across multiple benchmark datasets, suggesting prediction accuracy can be improved while maintaining efficiency by using the smaller model for most ``easy'' examples and the larger model for a small number of ``hard'' examples. 
As shown in Table \ref{tab:oracle-prompt}, assuming each data point is labeled as ``easy'' or ``hard'', this adaptive use of smaller and larger safety guard models improves the F1 score by 13\% and 10\% compared to using only the smaller or larger model on the WildGuardMix test split~\citep{wildguard}, respectively, while processing only 5.09\% of the dataset with the larger model.

\looseness=-1
Building on this key observation, we propose \textbf{SafeRoute}, a binary safety guard router designed to distinguish hard examples from easy ones. Given a dataset, we first label each instance as 1 if the smaller safety guard provides an incorrect prediction while the larger one provides an accurate prediction, as shown in~\Cref{tab:example}. Otherwise, we label it as 0.
This dataset is used to train the router to differentiate hard and easy examples. After training, the router classifies test instances into either category, deploying the smaller safety guard model for easy examples and the larger model for hard examples, as illustrated in~\Cref{fig:concept}.

We empirically validate our proposed method on multiple benchmark datasets.
Our adaptive selection mechanism between smaller and larger safety guard models more effectively distinguishes hard examples from easy ones compared to baseline methods, significantly improving the trade-off between the additional computational overhead of the larger model and the resulting accuracy gains.
Moreover, SafeRoute performs well not only on in-distribution (ID) data but also on out-of-distribution (OOD) scenarios, demonstrating its robustness across varying data distributions.

Our contributions and findings are summarized as follows:
\begin{itemize}
[itemsep=1mm,parsep=1pt,topsep=2pt,leftmargin=*]

\item We observe that some examples are easy, with the smaller safety guard model making correct predictions, while others are hard, with the smaller model failing but the larger safety guard model providing accurate predictions.

\looseness=-1
\item Based on this observation, we propose training a binary safety guard router, SafeRoute, to distinguish hard examples from easy ones. Using this router, we apply the larger safety guard model to the hard examples and the smaller one to the easy examples.

\item We empirically validate that our SafeRoute approach significantly improves the trade-off between accuracy gains and the additional overhead of using the larger model, across both ID and OOD datasets, compared to relevant baselines.

\end{itemize}
\section{Related Work}
\vspace{-0.05in}
\paragraph{Safety guard models.} Detecting harmful sentences has been a longstanding interest in the safety research community. Deep neural networks have been widely adopted to detect harmful user queries~\citep{hatebert, offensive-reddit, roberta-hate-speech}. Recently, LLMs with safety alignment have been prompted to judge the harmfulness of conversations between users and AI assistants~\citep{chao2024jailbreakbench}. 
Instead of relying on general-purpose LLMs, specialized safety guardrails are implemented by fine-tuning LLMs on labeled datasets~\citep{granite-guardian, wildguard, harmaug, metallamaguard3}. 
They moderate input prompts and output responses, thereby enabling the safe use of LLMs. 

\vspace{-0.05in}
\paragraph{Efficiency.} Deploying safety guard models alongside LLMs introduces additional computational overhead. To mitigate this cost, larger safety guard models are distilled into smaller ones~\citep{metallamaguard3, harmaug}. While this improves efficiency, smaller models typically underperform compared to their larger counterparts.
In this work, we aim to optimize the trade-off between computational overhead and accuracy by adaptively selecting between a larger and a smaller safety guard model based on input difficulty. Our approach is conceptually similar to speculative decoding~\citep{speculative-decoding, speculative-decoding-2, speculative-decoding-3}, where a smaller model generates a draft and a larger model verifies it, as both methods leverage models of different sizes to enhance computational efficiency.
Our method adaptively selects the model for each data point, allowing the larger model to be bypassed when appropriate. 
In contrast, speculative decoding always relies on the larger model to verify the smaller model’s output.

\section{Method}
\label{sec:method}
\subsection{Preliminaries}
\label{sec:preliminaries}
Given a user prompt $\rvx \in \mathcal{X}$ and its response $\rvy\in\mathcal{Y}$, generated by an LLM, we utilize a safety guard model $p: \mathcal{X} \times \mathcal{Y} \to [0,1]$ to predict its harmfulness, where $\mathcal{X}$ is the set of all possible prompts and $\mathcal{Y}$ is the set of all possible responses, including an empty response. The safety guard model estimates the probability of the pair being harmful as $p(c=1\mid \rvx, \rvy)$ and classifies it as harmful if the probability exceeds a threshold $\delta\in (0,1)$. Here $c\in\{0,1\}$ is a binary variable indicating the harmfulness of the prompt-response pair. Note that when the response  $\rvy$ is empty, the safety guard model only evaluates the harmfulness of the prompt $\rvx$.

\subsection{SafeRoute: Adaptive Model Selection}
\label{sec:adaptive_safety_guard_model_selection}
In this section, we introduce \textbf{SafeRoute}, our proposed adaptive mechanism for selecting safety guard models to optimize the trade-off between efficiency and accuracy.
\paragraph{Observation.}  We observe that a smaller safety guard model $q: \mathcal{X} \times\mathcal{Y}\to[0,1]$ correctly predicts harmfulness of many prompt-response pairs. 
However, there are cases where the larger safety guard model $p$ correctly classifies harmfulness, while the smaller safety guard model $q$ makes mistakes. 
Based on this, if we can identify which model makes the correct prediction for each prompt-response pair $(\rvx_i, \rvy_i)$, with label $c_i$, we can potentially improve prediction accuracy by selecting the appropriate safety guard model's prediction, while simultaneously minimizing the overhead of using the larger model, as follows: 
\begin{align*}
    \begin{cases}
        \one_{\{p(c=1 \mid \rvx_i, \rvy_i) > \delta \}}\text{,} 
        & \text{if } 
        \begin{array}[t]{rl}
            \one_{\{p(c=1 \mid \rvx_i, \rvy_i) > \delta\}} & = c_i, \\
            \one_{\{q(c=1 \mid \rvx_i, \rvy_i) > \delta\}} & \neq c_i
        \end{array} \\
        \one_{\{q(c=1 \mid \rvx_i, \rvy_i) > \delta \}}\text{,} 
        & \text{otherwise},
    \end{cases}
\end{align*}
where $\one$ denotes an indicator function. We use the prediction of the larger safety guard model, $p$, if it correctly classifies the prompt-response pair $(\rvx_i, \rvy_i)$, while the smaller model does not. Otherwise, we rely on the prediction of the smaller safety guard model, as there is no benefit to using the larger model in such cases.  

As shown in \Cref{tab:oracle-prompt}, this hypothetical combination of two safety guard models, denoted as ``Oracle'', achieves a significantly higher F1 score on the WildguardMix~\citep{wildguard} test split compared to using either the smaller model $q$, \texttt{Llama-Guard-3-1B}~\citep{metallamaguard3} or the larger model $p$, \texttt{Llama-Guard-3-8B}~\citep{metallamaguard3} alone, while utilizing only a small portion of the larger model.
\begin{table}[t]
    \centering
    \resizebox{0.45\textwidth}{!}{
    \begin{tabular}{llcc}
    \toprule    
    \textbf{Model} & \textbf{Type} & \textbf{F1} &  \textbf{Usage of Large} \\
    \midrule
    \rowcolor{rowgray}
    \texttt{Llama-Guard-3-1B}     & Small & 0.6702   & \phantom{00}0.00\% \\
    \texttt{Llama-Guard-3-8B} & Large & 0.7054 & 100.00\% \\
    \rowcolor{rowgray}
    Oracle  & Hybrid    &  \textbf{0.8101} & \phantom{00}5.09\% \\
    \bottomrule
    \end{tabular}
}
\vspace{-0.5em}
\caption{\looseness=-1
\textbf{Safety F1 score and larger model usage ratio} on the WildGuardMix test split~\cite{wildguard}.}
\label{tab:oracle-prompt}
\vspace{-0.15in}
\end{table}

\paragraph{Dataset creation and training.}
Building on the observation that some examples are ``easy'' while others are ``hard'', we propose training a binary safety guard router, \textbf{SafeRoute}, to distinguish between these instances. 
This allows for adaptive selection between smaller and larger safety guard models, thereby optimizing the trade-off between efficiency and accuracy compared to using either model in isolation.
To train \textbf{SafeRoute},  we use a dataset of prompt-response pairs with harmfulness labels, $\mathcal{D}=\{(\rvx_i, \rvy_i, c_i)  \}_{i=1}^n$, and assign a binary label $t_i\in\{0,1\}$ to each prompt-response pair, $(\rvx_i, \rvy_i)$, as follows:
\begin{align}
    t_i=\begin{cases}
        1\text{,} 
        & \text{if } 
        \begin{aligned}[t]
            \one_{\{p(c=1 \mid \rvx_i, \rvy_i) > \delta\}} &= c_i \\
            \text{and } \one_{\{q(c =1 \mid \rvx_i, \rvy_i) > \delta\}} &\neq c_i
        \end{aligned} \\
        0\text{,} 
        & \text{otherwise}.
    \end{cases}
\label{eq:label}
\end{align}
Then, we train a neural network-based router $f_\theta: \mathcal{X}\times \mathcal{Y}\to [0,1]$ to minimize the following binary cross-entropy loss:
\begin{align*}
\begin{split}
    \mathcal{L}(\theta;\hat{\mathcal{D}}) = -\frac{1}{\lvert \hat{\mathcal{D}}\rvert}&\sum_{(\rvx,\rvy, t)\in \hat{\mathcal{D}}} \big( t \cdot \log f_\theta(\rvx, \rvy) + \\ 
    &(1-t)  \cdot \log \left(1-f_\theta(\rvx, \rvy)\right) \big),
\end{split}
\end{align*}
where $\hat{\mathcal{D}}=\{(\rvx_i, \rvy_i, t_i)\}_{i=1}^n$.

\looseness=-1
\paragraph{Data augmentation.} Since the dataset $\hat{\mathcal{D}}$ contains only a small number of examples with label $t_i=1$,  we augment the training dataset $\mathcal{D}$ with paraphrased inputs. Specifically, we prompt the LLM, \texttt{Llama-3.1-8B-Instruct}~\citep{llama-3}, to generate multiple paraphrases for each prompt-response pair $(\rvx,\rvy)\in\mathcal{D}$. We then label both the synthesized dataset and the original dataset following~\Cref{eq:label}, resulting in an augmented dataset $\hat{\mathcal{D}}_\texttt{aug}=\{(\rvx_i, \rvy_i, t_i) \}_{i=1}^m$. Finally, we train the router $f_\theta$ to minimize the loss $\mathcal{L}(\theta;\hat{\mathcal{D}}_\texttt{aug})$.

\paragraph{Parameterization.} There are many ways to parameterize the binary router $f_\theta$. However, additional overhead of utilizing $f_\theta$ should be minimized to ensure efficiency. Moreover, for better decision-making, the router should capture what the smaller safety guard model, $q$, knows and does not know about its input. To achieve this, we extract the last token’s hidden representation from the final layer of the smaller safety guard model, as the safety guard model directly uses this last token representation for harmfulness prediction. The binary router can utilize this extracted feature to learn patterns of correct and incorrect predictions. For efficient training and inference, we always freeze the feature extractor, which enables us to reuse the last layer feature for predictions of harmfulness with $q$.

\paragraph{Inference.} At inference time, for given a test prompt-response pair $(\rvx_*, \rvy_*)$, we compute the score of selecting the larger model as $f_\theta(\rvx_*, \rvy_*)$. If the score exceeds a certain threshold $\epsilon \in (0,1)$, we utilize the larger safety guard model $p$ to predict the harmfulness of the prompt-response pair $(\rvx_*, \rvy_*)$. Otherwise, we use the smaller safety guard model $q$ for the prediction of $(\rvx_*, \rvy_*)$.

\subsection{Theoretical analysis} To further understand the effectiveness of our proposed adaptive approach, we provide a theoretical analysis of its risk bound. Specifically, we analyze how the selection mechanism, governed by the router $f_\theta$, influences the overall performance by comparing the risk of the adaptive model to that of an oracle model with perfect selection.

Let $\ell(p(\rvx,\rvy), c)= -(c \log p(c=1\mid \rvx,\rvy) +(1-c)\log p(c=0\mid \rvx,\rvy))$ be the binary cross-entropy loss with the larger safety guard model $p$ and labeled data $(\rvx,\rvy,c)$. The loss $\ell(q(\rvx,\rvy),c)$ is defined in the same manner for $q$. We define, $I(\rvx,\rvy) = \one_{\{ f_\theta(\rvx,\rvy) >\epsilon \}}$, where  the router $f_\theta$ determines which safety guard model is selected. The risk of our adaptive model given $p$ and $q$ is: 
\begin{align*}
\begin{split}
    R_\text{adaptive} = \mathbb{E}&[I(\rvx,\rvy)\ell(p(\rvx,\rvy), c)  \\+
    &(1-I(\rvx,\rvy))\ell(q(\rvx,\rvy), c) ],
\end{split}
\end{align*}
where the expectation is taken over an unknown data distribution.  The oracle risk is then given by:
\begin{align*}
    \begin{split}
        R_\text{oracle} = \mathbb{E}&[t(\rvx,\rvy)\ell(p(\rvx,\rvy),c) 
        \\+ &(1-t(\rvx,\rvy)) \ell(q(\rvx,\rvy),c)],
    \end{split}
\end{align*}
where $t(\rvx,\rvy)$ represents the optimal model selection strategy, as defined in~\Cref{eq:label}. 
\looseness=-1
\begin{thm}
Assuming that $\mathbb{E}[\lvert \ell(p(\rvx,\rvy),c)-\ell(q(\rvx,\rvy),c)\rvert^2]$ is bounded, we can bound the risk of our adaptive model as follows:
\begin{equation*}
\begin{gathered}
    R_\text{adaptive} \leq R_\text{oracle} + M\sqrt{\mathbb{P}\left(I(\rvx,\rvy)\neq t(\rvx,\rvy)\right)},
\end{gathered}
\end{equation*}
where $M=\sqrt{\mathbb{E}[\lvert \ell(p(\rvx,\rvy),c)-\ell(q(\rvx,\rvy),c)\rvert^2]}$.
\label{thm}
\end{thm}
The proof is deferred to~\Cref{app:proof}.
This theorem indicates that the gap between $R_{\text{adaptive}}$ and $R_{\text{oracle}}$ depends on the probability of incorrect selection $\mathbb{P}(I(\rvx,\rvy)\neq t(\rvx,\rvy))$, which decreases as $f_\theta$ improves. Consequently, as the number of training samples for $f_\theta$ increases, reducing its generalization error, the risk bound tightens. In the asymptotic case where $f_\theta$ perfectly approximates $t$, we achieve $R_{\text{adaptive}} = R_{\text{oracle}}$. In contrast, other entropy-based model selection baselines, described in~\Cref{sec:exp}, do not guarantee such optimality. 
A smaller model, even with perfect calibration, cannot predict what the larger model knows and therefore cannot reduce the error. 

\section{Experiments}

\subsection{Experimental Setups}
\label{sec:exp}
\paragraph{Datasets.} For the training dataset $\mathcal{D}$, we use the train split of WildGuardMix~\citep{wildguard}. We evaluate our method on six public benchmark datasets: the test split of \textbf{WildGuardMix}, \textbf{WildGuardMix-p}, 
OpenAI Moderation~\citep[\textbf{OAI};][]{oai-ds}, \textbf{ToxicChat}~\citep{toxic-chat}, \textbf{XSTest}~\cite{xstest}, and \textbf{HarmBench}~\citep{harmbench}. The WildGuardMix-p dataset is a subset of the WildGuardMix test split, containing only instances with prompt harmfulness labels, excluding those without them.
WildGuardMix-p, OAI, and ToxicChat datasets are used for prompt classification (\ie, a response is always an empty sequence), while the others are for prompt-response pair classification. Please see \Cref{tab:data-stat} in \Cref{sec:data_statistics} for data statistics.

\looseness=-1
\paragraph{Implementation details.} 
We use \texttt{Llama-Guard} \\ \texttt{-3-1B}~\citep{metallamaguard3} as the smaller model \( q \) and \texttt{Llama-Guard-3-8B}~\citep{metallamaguard3} or \texttt{Granite-Guardian-3-8B}~\citep{granite-guardian} as the larger model \( p \). Following \citet{liu2024calibration}, we define the safety binary distribution as follows:
\begin{align*}
p(c=1|\rvx,\rvy) = \frac{\exp(z_{p,1})}{\exp(z_{p,0}) + \exp(z_{p,1})},
\end{align*}
where \( z_{p,0} \) and \( z_{p,1} \) are the logits of the safe and unsafe tokens from the safety guard model $p$. We use 10\% of the WildGuardMix training split as a validation set for tuning \( f_\theta \) and set the number of paraphrases per example to 7. The input features of \( f_\theta \) are the last-layer outputs of the small model, selecting only the final token.
We implement \( f_\theta \) as a three-layer Bayesian neural network~\citep{blundell2015weight}, where each layer consists of an affine transformation, layer normalization~\cite{ba2016layer}, and a ReLU~\citep{Nair2010RectifiedLU} activation, except in the last layer. The posterior is approximated by a Gaussian with a diagonal covariance matrix, while the prior follows \( \mathcal{N}(0, 0.1) \). The Kullback-Leibler divergence weight is set to 0.01.
To maintain efficiency, we use 1 Monte Carlo sample for both training and inference. We train \( f_\theta \) for 1000 epochs with a mini-batch size of 512, approximately balancing \( t = 0 \) and \( t = 1 \) per batch. The parameters \( \theta \) are optimized using Adam~\cite{kingma2014adam} with a 0.001 learning rate, linear decay, and 100 warmup steps.
We run experiments five times with different random seeds for the \textbf{Random} baseline and \textbf{SafeRoute}, both of which involve stochastic components. All experiments are conducted on a single \href{https://www.nvidia.com/en-us/data-center/h200/}{NVIDIA H200 Tensor Core GPU}.
We present the \href{https://huggingface.co/docs/hub/index}{Hugging Face Hub} identifiers for all pretrained models used in this paper in \Cref{tab:model} of \Cref{sec:model}.

\begin{table*}[t]
\centering
\resizebox{0.98\textwidth}{!}{

\begin{tabular}{lccccccc}

\toprule
\multirow{2}{*}{\textbf{Method}} & \multicolumn{3}{c}{\textbf{Prompt-only}} & \multicolumn{3}{c}{\textbf{Prompt-Response}} & \multirow{2}{*}{\textbf{Average}} \\
\cmidrule(lr){2-4}
\cmidrule(lr){5-7}
& {\textbf{WildGuardMix-p}}
& {\textbf{ToxicChat}}
& {\textbf{OAI}}
& {\textbf{WildGuardMix}}
& {\textbf{XSTest}}
& {\textbf{HarmBench}}
&  \\

\midrule
\rowcolor{rowgray}
Entropy & \underline{0.3110} & \underline{0.4002} & \bf0.4174 & \underline{0.2947} & \underline{0.2466} & \underline{0.4094} & \underline{0.3465} \\
+TS     & 0.1641 & 0.2004 & 0.2626 & 0.1046 & 0.0680 & 0.1930 & 0.1655 \\
\rowcolor{rowgray}
+CC     & 0.2852 & 0.3135 & 0.3472 & 0.2470 & 0.1978 & 0.3786 & 0.2949 \\
+BC     & 0.2264 & 0.1854 & 0.2098 & 0.1433 & 0.1228 & 0.3262 & 0.2023 \\
\midrule
\rowcolor{rowgray}
\textbf{SafeRoute (Ours)}    & \textbf{0.5054}\tiny$\pm$0.0098 & \textbf{0.5682}\tiny$\pm$0.0103 & \underline{0.3501}\tiny$\pm$0.0170 & \textbf{0.5434}\tiny$\pm$0.0153 & \textbf{0.4991}\tiny$\pm$0.0297 & \textbf{0.5124}\tiny$\pm$0.0086 & \textbf{0.4964}\tiny$\pm$0.0111 
\\
\bottomrule
\end{tabular}
}
\vspace{-0.5em}
\caption{
\textbf{Routing F1 score} using the smaller (\texttt{Llama-Guard-3-1B}) and larger (\texttt{Llama-Guard-3-8B}) models.
The best results are in \textbf{bold}, and the second-best ones are \underline{underlined}.
} 
\label{tab:binary_classification}
\vspace{-0.1in}
\end{table*}

\begin{figure*}[t]
\centering
\includegraphics[width=0.99\textwidth]{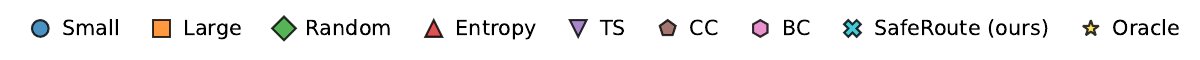}
\medskip
\vspace{-0.15in}
\includegraphics[width=0.99\textwidth]{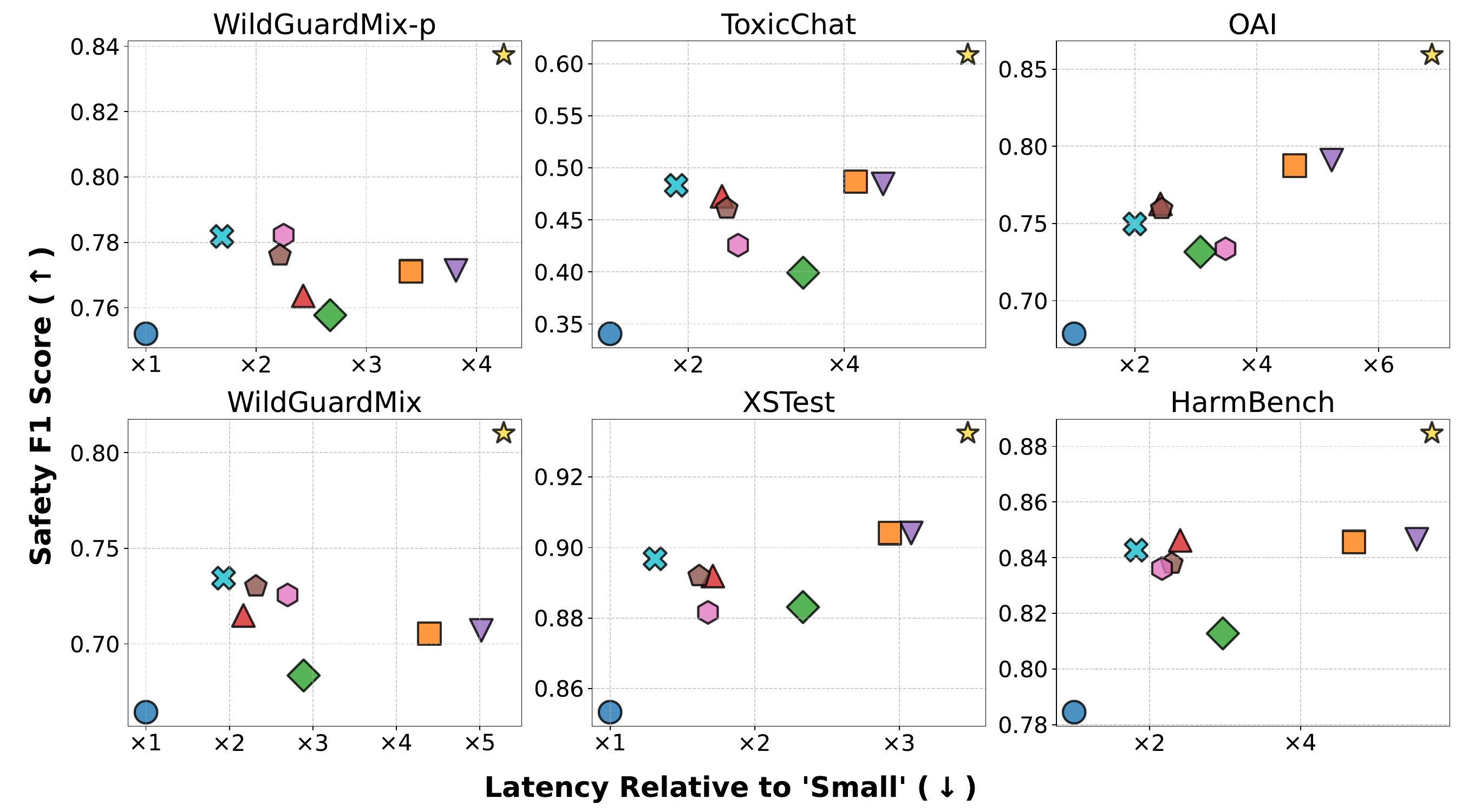}
\vspace{-0.5em}
\caption{\looseness=-1
\textbf{Latency ($\downarrow$) vs. safety F1 score ($\uparrow$) trade-off} when using the smaller (\texttt{Llama-Guard-3-1B}) and larger (\texttt{Llama-Guard-3-8B}) models. See \Cref{fig:flops} and \ref{fig:large_ratio} in \Cref{sec:additional_experimental_results} for FLOPs and ratio of large model trade-off.}
\label{fig:latency}
\vspace{-0.15in}
\end{figure*}

\paragraph{Baselines.} We compare our method against the following baselines: 
\begin{enumerate}[leftmargin=0in, topsep=5pt, leftmargin=*]
\item[1-2.] \textbf{Small} and \textbf{Large}: These methods use either only the smaller or larger safety guard models.

\vspace{-0.05in}
\item[3.] \textbf{Random}: This method randomly selects a safety guard model, choosing the larger one with 50\% probability.

\vspace{-0.05in}
\item[4.] \textbf{Entropy}: In this method, the entropy of smaller safety guard model is computed as follows:
\begin{align*}
    \begin{split}
    H(\rvx,\rvy)=&-q(c=0|\rvx,\rvy) \log_2 q(c=0|\rvx,\rvy) \\ 
    &-q(c=1|\rvx,\rvy) \log_2 q(c=1|\rvx,\rvy).
    \end{split}
\end{align*}
When the entropy exceeds 0.5, indicating high uncertainty, we use the larger safety guard model. In the following three calibration methods (\textbf{TS}, \textbf{CC}, and \textbf{BC}), we calibrate the distribution $q$ of the smaller guard model to improve uncertainty estimation for better decision-making.

\vspace{-0.05in}
\item[5.] \textbf{Temperature Scaling (TS)}~\cite{guo2017calibration}: This method is a widely used confidence calibration technique for neural networks. We divide the logits, $z_{q,0}$ and $z_{q,1}$, of the smaller safety guard model $q$  by $\tau\in\mathbb{R}_{>0}$ and renormalize it:
\begin{equation*}
    \hat{q}(c=1\mid \rvx,\rvy)= \frac{\exp(z_{q,1}/\tau)}{\exp(z_{q,0}/\tau) + \exp(z_{q,1}/\tau)}.
\end{equation*}
We optimize $\tau$ to maximize the log-likelihood of the  WildGuardMix training split~\citep{wildguard}. Then we 
compute the entropy $H(\rvx,\rvy)$ using the calibrated distribution $\hat{q}$ and select the larger model if the entropy exceeds 0.5; otherwise, the smaller model is chosen.

\vspace{-0.05in}
\item[6.] \textbf{Contextual Calibration (CC)}~\cite{zhao2021calibrate}: This method is a matrix scaling technique designed to mitigate contextual bias in LLMs, with the key advantage of not requiring a validation set. It calibrates the output distribution of $q$ using content-free tokens, such as a string of whitespace, $\emptyset=\text{`` ''}$, as follows:
\begin{align*}
\hat{q}(c=1|\rvx,\rvy) &= \frac{\frac{q(c=1|\rvx,\rvy)}{q(c=1|\emptyset)}}{\frac{q(c=0|\rvx,\rvy)}{q(c=0|\emptyset)}+\frac{q(c=1|\rvx,\rvy)}{p(c=1|\emptyset)}}
\end{align*}
with $\hat{q}(c=0\mid \rvx, \rvy)=1-\hat{q}(c=1\mid \rvx, \rvy)$. Similar to TS, we select the larger model $p$ based on the entropy with the distribution $\hat{q}$.

\vspace{-0.05in}
\item[7.] \textbf{Batch Calibration (BC)}~\cite{zhou2023batch}: 
BC is another matrix scaling technique that calibrates the output distribution $q$ using batch probabilities ($\bar{q}_0, \bar{q}_1$), computed as follows:
\begin{align*}
\hat{q}(c=1|\rvx,\rvy) &= \frac{\frac{q(c=1|\rvx, \rvy)}{\bar{q}_1}}{\frac{q(c=0|\rvx, \rvy)}{\bar{q}_0}+\frac{q(c=1|\rvx, \rvy)}{\bar{q}_1}}
\end{align*}
with $\hat{q}(c=0\mid \rvx, \rvy)=1-\hat{q}(c=1\mid \rvx, \rvy)$, where $\bar{q}_1 = \frac{1}{|\mathcal{D}^\prime|} \sum_{(\rvx^\prime,\rvy^\prime)\in\mathcal{D}^\prime} q(c=1|\rvx', \rvy')$ and $\bar{q}_0 = 1-\bar{q}_1$. 
For a fair comparison, we use the training split of WildGuardMix for $\mathcal{D}^\prime$ (\ie, $\mathcal{D}^\prime = \mathcal{D}$). Similar to TS, we select the larger safety guard model based on the entropy with the distribution $\hat{q}$.

\vspace{-0.05in}
\item[8.] \textbf{Oracle}: As described in \Cref{sec:adaptive_safety_guard_model_selection}, this method combines the smaller and larger safety guard models, using the larger one only when the smaller one is incorrect and the larger one is correct. Assuming access to the true label $c$, it provides an upper bound on accuracy for adaptive model selection. However, it always requires two forward passes, one for the smaller model and one for the larger model, making it the most computationally expensive method.

\end{enumerate}


\begin{table*}[t]
\centering
\resizebox{0.98\textwidth}{!}{

\begin{tabular}{lccccccc}

\toprule
\multirow{2}{*}{\textbf{Method}} & \multicolumn{3}{c}{\textbf{Prompt-only}} & \multicolumn{3}{c}{\textbf{Prompt-Response}} & \multirow{2}{*}{\textbf{Average}} \\
\cmidrule(lr){2-4}
\cmidrule(lr){5-7}
& {\textbf{WildGuardMix-p}}
& {\textbf{ToxicChat}}
& {\textbf{OAI}}
& {\textbf{WildGuardMix}}
& {\textbf{XSTest}}
& {\textbf{HarmBench}}
&  \\

\midrule
\rowcolor{rowgray}
Entropy & {0.4059} & \underline{0.3899} & \bf0.3639 & \underline{0.3176} & \underline{0.2778} & \underline{0.4345} & \underline{0.3649} \\
+TS     & 0.2868 & 0.2277 & 0.2591 & 0.1274 & 0.0625 & 0.2264 & 0.1983 \\
\rowcolor{rowgray}
+CC     & 0.4254 & 0.3125 & 0.3191 & 0.2620 & 0.2222 & 0.3828 & 0.3207 \\
+BC     & \underline{0.4373} & 0.2064 & 0.2232 & 0.1776 & 0.1239 & 0.2846 & 0.2422 \\
\midrule
\rowcolor{rowgray}
\textbf{SafeRoute (Ours) } & \textbf{0.6128}\tiny$\pm$0.0059 & \textbf{0.4887}\tiny$\pm$0.0114 & \underline{0.3257}\tiny$\pm$0.0044 & \textbf{0.6141}\tiny$\pm$0.0124 & \textbf{0.5621}\tiny$\pm$0.0297 & \textbf{0.5592}\tiny$\pm$0.0173 & \textbf{0.5271}\tiny$\pm$0.0053 \\
\bottomrule
\end{tabular}
}
\vspace{-0.5em}
\caption{
\textbf{Routing F1 score} using the smaller (\texttt{Llama-Guard-3-1B}) and larger (\texttt{Granite-Guardian-3-8B}) models.
The best results are in \textbf{bold}, and the second-best ones are \underline{underlined}.} 
\label{tab:binary_classification_guardian}
\vspace{-0.1in}
\end{table*}

\begin{figure*}[t]
\centering
\includegraphics[width=0.99\textwidth]{images/legend.pdf}
\medskip
\vspace{-0.15in}
\includegraphics[width=0.99\textwidth]{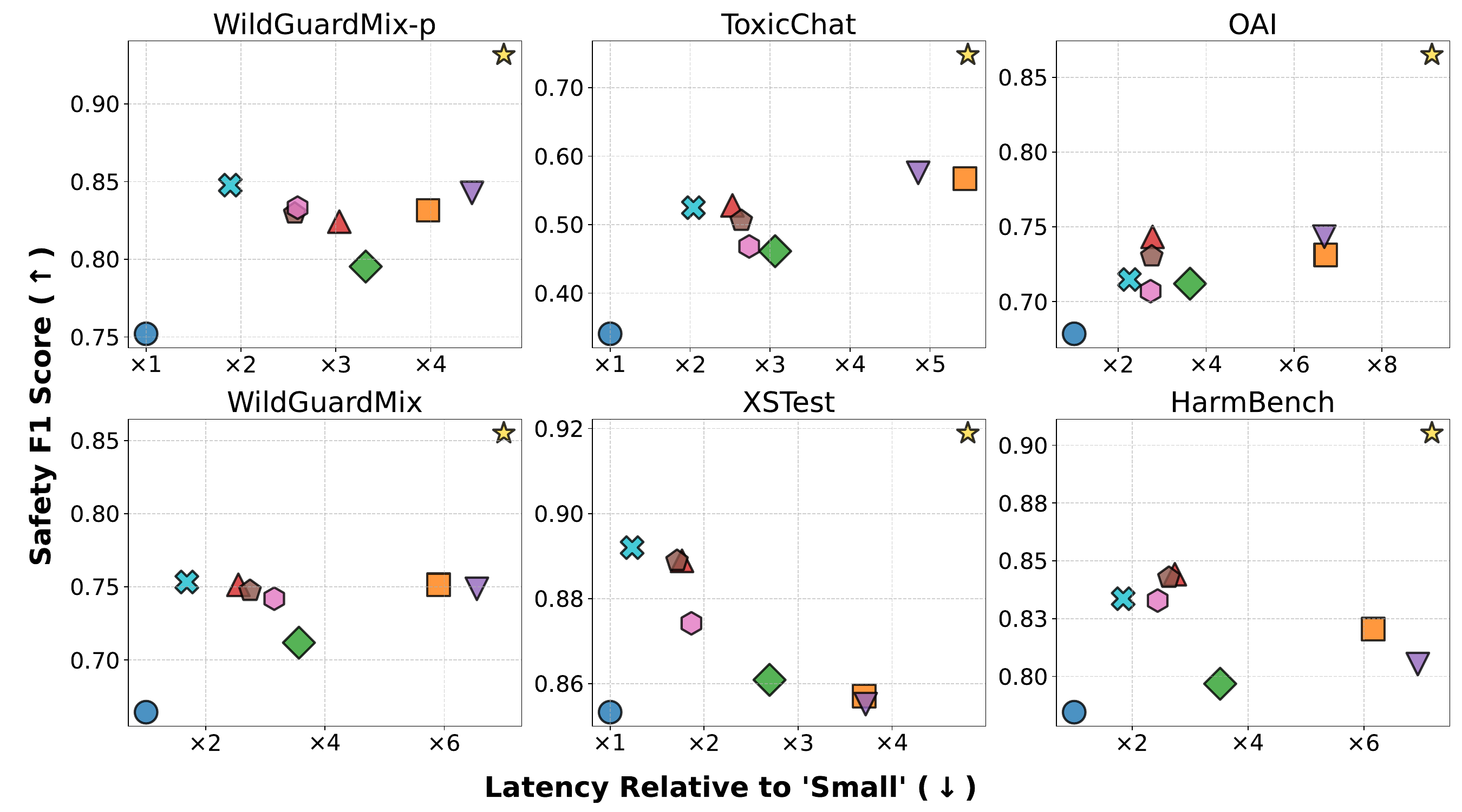}
\vspace{-0.5em}
\caption{\looseness=-1
\textbf{Latency ($\downarrow$) vs. safety F1 score ($\uparrow$) trade-off} when using the smaller (\texttt{Llama-Guard-3-1B}) and larger (\texttt{Granite-Guardian-3-8B}) models. See \Cref{fig:flops_guardian} and \ref{fig:large_ratio_guardian} in \Cref{sec:additional_experimental_results} for FLOPs and ratio of large model trade-off.}
\label{fig:latency_guardian}
\vspace{-0.15in}
\end{figure*}
\subsection{Experimental results.}

\paragraph{Routing results using \texttt{Llama-Guard-3-8B}.} To evaluate how accurately our SafeRoute model $f_\theta$ is able to distinguish hard examples from easy ones, we compare its routing predictions with the corresponding labels $t_i$, as defined in~\Cref{eq:label}, and compute F1 score. As shown in~\Cref{tab:binary_classification}, SafeRoute outperforms naive entropy-based methods, such as TS, CC, and BC, by a large margin on most benchmark datasets, except for OAI. The performance of SafeRoute shows the importance of learning to identify examples where the larger model classifies correctly while the smaller model makes errors. While the entropy of the smaller model correlates with its likelihood of making incorrect predictions, it provides no insight into the behavior of the larger model. This limitation leads to an increased number of false positives, resulting in lower F1 scores compared to our approach. 


\begin{figure*}[ht]
    \centering
    \includegraphics[width=0.7\textwidth]{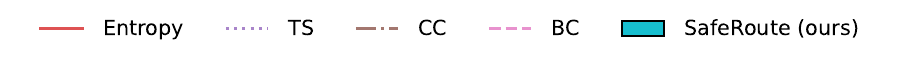}
    \medskip    
    \begin{subcaptionbox}{\label{fig:pooling_ablation}}%
        {\includegraphics[width=0.32\textwidth]{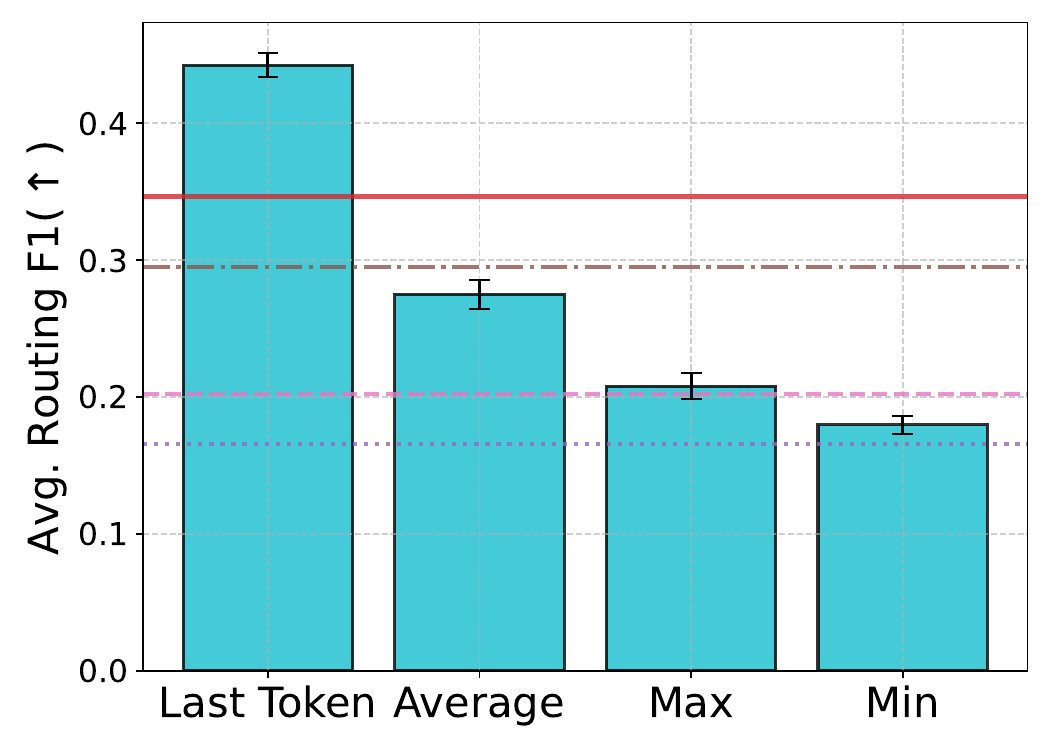}\vspace{-0.05in}}
    \end{subcaptionbox}
    \hfill
    \begin{subcaptionbox}{\label{fig:layer_ablation}}%
        {\includegraphics[width=0.32\textwidth]{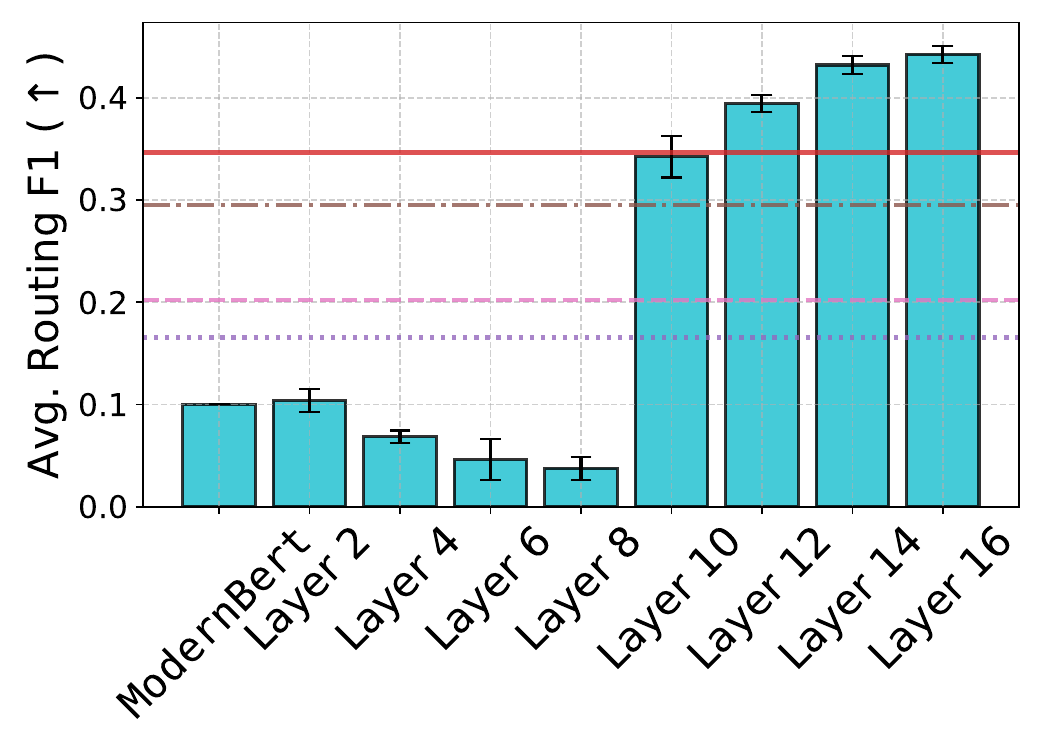}\vspace{-0.05in}}
    \end{subcaptionbox}
    \hfill
    \begin{subcaptionbox}{\label{fig:aug_ablation}}%
        {\includegraphics[width=0.32\textwidth]{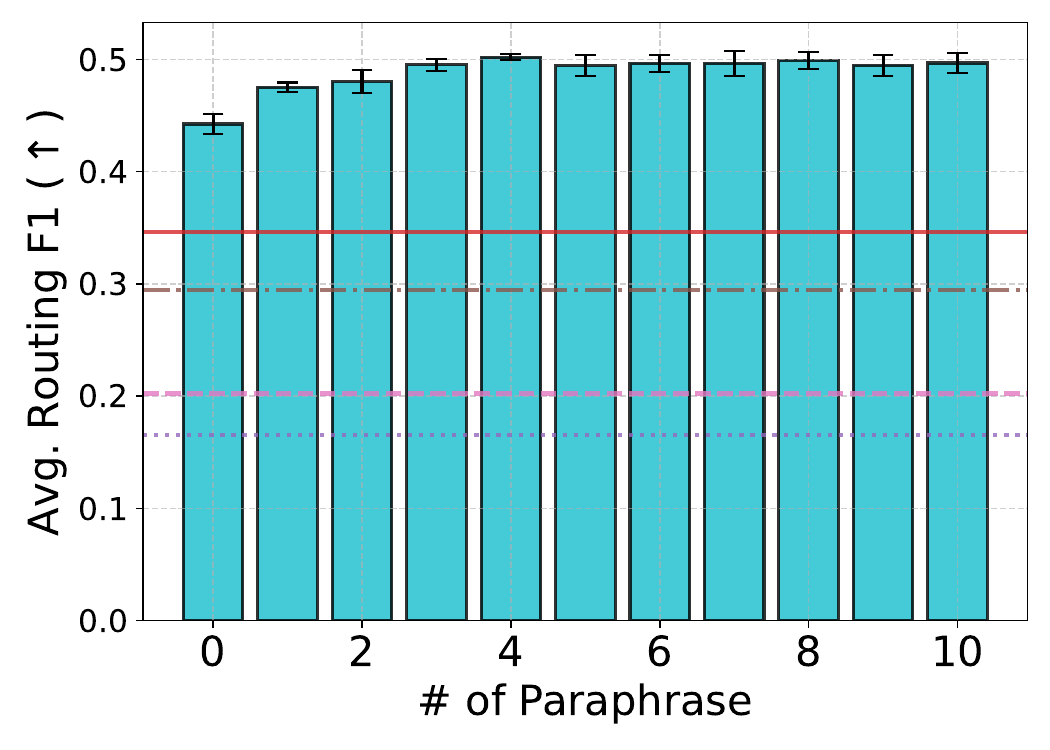}\vspace{-0.05in}}
    \end{subcaptionbox}
    \vspace{-0.1in}
    \caption{Ablation studies on \textbf{(a)}: pooling methods, \textbf{(b)}: feature layers, and \textbf{(c)}: the number of paraphrases.}
    \vspace{-0.15in}
\end{figure*}
\begin{figure}[t]
\medskip
\centering
\includegraphics[width=0.95\columnwidth]{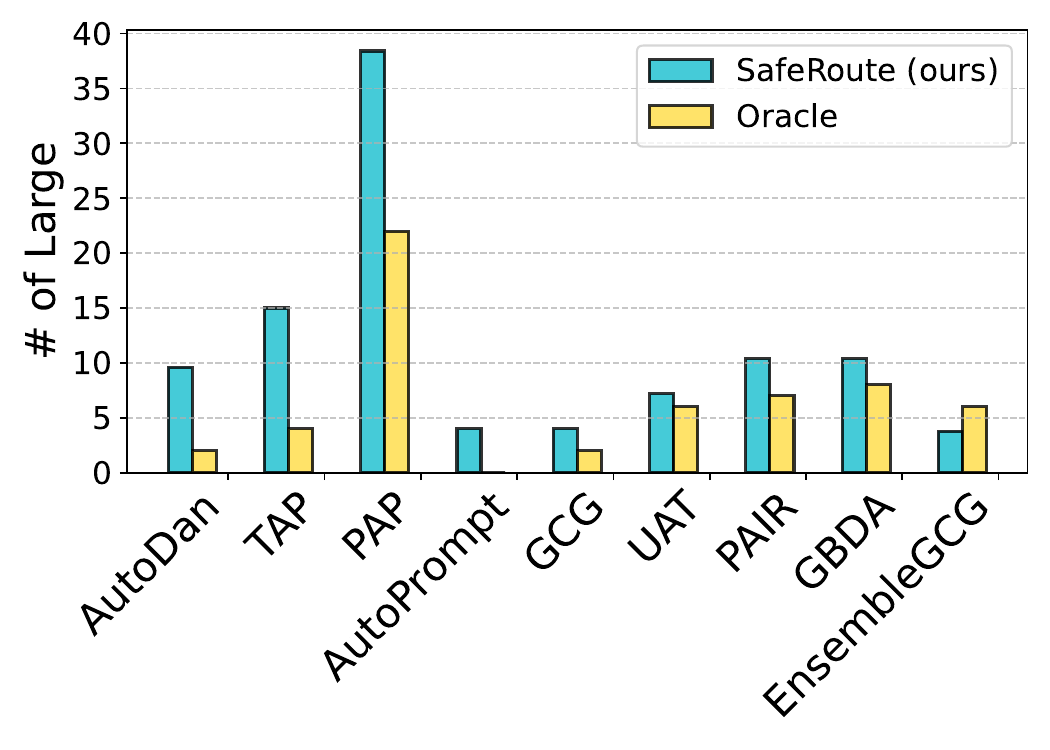}
\vspace{-0.5em}
\caption{The number of the large model selections for each jailbreak attack in HarmBench dataset.}
\label{fig:jailbreaktype}
\vspace{-0.15in}
\end{figure}

\paragraph{Trade-off using \texttt{Llama-Guard-3-8B}.}
We observe a similar pattern in trade-off between latency and F1 score when adaptively selecting between smaller and larger models. As shown in~\Cref{fig:latency}, SafeRoute significantly improves the F1 score over using the smaller model alone while achieving performance comparable to the larger model. Moreover, the increase in latency due to using the larger model on some examples is smaller than that of any baseline. 
This can be attributed SafeRoute's more accurate routing decisions compared to entropy-based methods, which frequently misclassify examples and introduce significantly higher computational overhead. 
We present the average of safety F1 score, precision, recall, and latency in \Cref{tab:full_8B}.

\vspace{-0.05in}
\paragraph{Routing results using \texttt{Granite-Guardian-3-8B}.}
In addition to \texttt{Llama-Guard-3-8B}, we train the router $f_\theta$ using \texttt{Llama-Guard-3-1B} and \texttt{Granite-Guardian-3-8B}, and evaluate the router on the same six benchmark datasets used in previous experiments. As shown in~\Cref{tab:binary_classification_guardian}, our proposed SafeRoute  more accurately distinguishes hard examples from easy ones across all datasets except for OAI, which is consistent with the results from previous experiments.

\vspace{-0.05in}
\paragraph{Trade-off using \texttt{Granite-Guardian-3-8B}.}
When \texttt{Granite-Guardian-3-8B} is used, the improved routing ability also leads to a better trade-off between latency and F1 score improvements compared to other baselines across four datasets, as illustrated in~\Cref{fig:latency_guardian}. For OAI and Harmbench, SafeRoute achieves lower latency but slightly lower F1 score gains than the CC and Entropy baselines. Although some entropy-based selection methods improve F1 score relative to using the smaller model alone, they introduce significantly higher latency overhead by more frequently selecting the larger model even when it provides no performance benefit.
We present the average of safety F1 score, precision, recall, and latency in \Cref{tab:full_guardian}.

\paragraph{Ablation studies.} 
We conduct ablation studies to evaluate how our design choices in \textbf{SafeRoute} affect performance, reporting the average routing F1 score across the six benchmark datasets used in previous experiments. Specifically, we examine the impact of:  
\textbf{(a)} Replacing the original sequence pooling method (last token) with an average, maximum, or minimum operator.  
\textbf{(b)} Replacing features from the smaller model $q$ with those from \texttt{ModernBERT}~\citep{modernbert}, a bidirectional encoder based on BERT~\citep{bert} with rotary positional embeddings~\citep{rope} and local-global alternating attention. We also explore using features from layers of the smaller model other than the last (16th) layer.  
\textbf{(c)} Removing paraphrased prompt-response pairs from the training dataset $\mathcal{D}$.  

As shown in \Cref{fig:pooling_ablation}, using the last token as the feature for our router $f_\theta$ improves the average routing F1 score across all six datasets, highlighting both the simplicity and effectiveness of using the last token.  
\Cref{fig:layer_ablation} shows the importance of how inputs to the router is encoded. Notably, replacing features from the smaller model $q$ with \texttt{ModernBERT} features leads to severe overfitting, suggesting that \texttt{ModernBERT} fails to capture the uncertainties of $q$ and does not generalize well to unseen examples. This highlights the importance of leveraging features from the smaller model rather than relying on an external encoder. Additionally, using features from layers other than the last layer results in underperformance, indicating that these layers do not accurately capture what the smaller model does not know.  
Finally, as seen in \Cref{fig:aug_ablation}, removing paraphrased data degrades generalization performance, while increasing the number of paraphrases per example improves performance. However, performance plateaus beyond a certain number of paraphrases, likely due to limited diversity. Developing methods to synthesize diverse, high-quality data for augmentation remains an interesting direction for future work.

\paragraph{Analysis of jailbreak attacks.} We analyze how SafeRoute selects the larger safety guard model for different jailbreak attacks in the HarmBench dataset. Specifically, we examine its behavior against AutoDan~\citep{autodan}, TAP~\citep{tap}, PAP~\citep{pap}, AutoPrompt~\citep{autoprompt}, GCG~\citep{zou2023universal}, UAT~\citep{uat}, PAIR~\citep{pair}, and GBDA~\citep{gbda}. As shown in~\Cref{fig:jailbreaktype}, both the oracle and SafeRoute select the larger model most frequently for the PAP attack. Since this attack exploits persuasive taxonomy to elicit harmful responses from LLMs, the smaller model is more prone to errors than other types of attacks. On the other hand, both models select the larger model less frequently for the GCG attack. This may be attributed to the fact that this jailbreak attack is well-known, and many of its instances are included in the dataset used to train the smaller model.

\section{Conclusion}
In this work, we proposed training a binary router, SafeRoute, that adaptively selects either a larger or smaller safety guard model based on the difficulty of the input data. This approach improved the trade-off between computational overhead and accuracy gains compared to other relevant baselines on several benchmark datasets. While we focused on the dynamic selection of safety guard models with different sizes, our approach is not limited to prompt-response pair classification. An interesting direction for future work is extending this method to other tasks, such as reasoning or programming.

\section*{Limitations}
Although our proposed adaptive selection between a smaller and a larger safety guard model significantly improves the trade-off between accuracy gains and computational overhead compared to other baselines, it has some limitations. First, the current parameterization of the binary classifier $f_\theta$ does not encode what the larger model knows, limiting its generalization performance. In our preliminary experiments, we incorporated representations of the larger model as part of the classifier’s input. While this improved accuracy, it introduced significant computational overhead, making the approach even slower than using the larger model alone. Approximating the larger model's features in an efficient manner would be an interesting direction as future work. Another limitation is that the performance of our selection mechanism is highly dependent on the quality and representativeness of the training data for $f_\theta$. If the training dataset does not adequately capture the diversity of prompt-response pairs --- particularly those at the boundary between easy and hard instances --- the classifier may make suboptimal decisions. Steering LLMs to generate diverse and high-quality data is another promising avenue for future work.

\section*{Ethics Statement}
Our proposed method, SafeRoute, aims to improve the trade-off between efficiency and accuracy gains of safety guard models in large language model (LLM) deployment. We do not foresee any direct ethical concerns arising from the use of SafeRoute, as it functions solely as an adaptive mechanism for selecting between smaller and larger models based on their predictive performance across different input types. By doing so, it ensures a more efficient deployment while maintaining high safety performance, reducing computational overhead without compromising the ability to detect harmful inputs. We are committed to the responsible use of LLMs and the enhancement of safety mechanisms, ensuring that no additional harm is introduced by our approach. All experiments were conducted with publicly available benchmark datasets. 

\section*{Acknowledgement}
This work was partially supported by the Institute of Information \& Communications Technology Planning \& Evaluation (IITP) grant funded by the Korea government (MSIT) (No. RS-2020-II200153, Penetration Security Testing of ML Model Vulnerabilities and Defense), 
Institute for Information \&
communications Technology Promotion (IITP) grant funded by the Korea government (MSIT)
(No. RS-2019-II190075, Artificial Intelligence Graduate School Program (KAIST)), 
Institute of Information \& communications Technology Planning \& Evaluation(IITP) grant funded by the Korea government(MSIT) (No.RS-2022-II220713, Meta-learning Applicable to Real-world Problems),
Samsung Electronics (IO201214-08145-01), 
and National Research Foundation of Korea (NRF) grant funded by the Korea government (MSIT) (No. RS-2023-00256259).

\bibliography{reference}

\begin{thebibliography}{41}
\providecommand{\natexlab}[1]{#1}

\bibitem[{Ba(2016)}]{ba2016layer}
Jimmy Ba. 2016.
\newblock \href {https://arxiv.org/abs/1607.06450} {Layer normalization}.
\newblock \emph{arXiv preprint arXiv:1607.06450}.

\bibitem[{Blundell et~al.(2015)Blundell, Cornebise, Kavukcuoglu, and Wierstra}]{blundell2015weight}
Charles Blundell, Julien Cornebise, Koray Kavukcuoglu, and Daan Wierstra. 2015.
\newblock \href {https://proceedings.mlr.press/v37/blundell15} {Weight uncertainty in neural network}.
\newblock \emph{International Conference on Machine Learning (ICML)}.

\bibitem[{Caselli et~al.(2021)Caselli, Basile, Mitrovi{\'c}, and Granitzer}]{hatebert}
Tommaso Caselli, Valerio Basile, Jelena Mitrovi{\'c}, and Michael Granitzer. 2021.
\newblock \href {https://doi.org/10.18653/v1/2021.woah-1.3} {{H}ate{BERT}: Retraining {BERT} for abusive language detection in {E}nglish}.
\newblock In \emph{Proceedings of the 5th Workshop on Online Abuse and Harms (WOAH 2021)}, pages 17--25, Online. Association for Computational Linguistics.

\bibitem[{Chao et~al.(2024)Chao, Debenedetti, Robey, Andriushchenko, Croce, Sehwag, Dobriban, Flammarion, Pappas, Tram{\`e}r, Hassani, and Wong}]{chao2024jailbreakbench}
Patrick Chao, Edoardo Debenedetti, Alexander Robey, Maksym Andriushchenko, Francesco Croce, Vikash Sehwag, Edgar Dobriban, Nicolas Flammarion, George~J. Pappas, Florian Tram{\`e}r, Hamed Hassani, and Eric Wong. 2024.
\newblock \href {https://openreview.net/forum?id=urjPCYZt0I} {Jailbreakbench: An open robustness benchmark for jailbreaking large language models}.
\newblock \emph{Advances in Neural Information Processing Systems (NeurIPS) Datasets and Benchmarks Track}.

\bibitem[{Chao et~al.(2023)Chao, Robey, Dobriban, Hassani, Pappas, and Wong}]{pair}
Patrick Chao, Alexander Robey, Edgar Dobriban, Hamed Hassani, George~J Pappas, and Eric Wong. 2023.
\newblock \href {https://arxiv.org/abs/2310.08419} {Jailbreaking black box large language models in twenty queries}.
\newblock \emph{arXiv preprint arXiv:2310.08419}.

\bibitem[{Chen et~al.(2023)Chen, Borgeaud, Irving, Lespiau, Sifre, and Jumper}]{speculative-decoding-2}
Charlie Chen, Sebastian Borgeaud, Geoffrey Irving, Jean-Baptiste Lespiau, Laurent Sifre, and John Jumper. 2023.
\newblock \href {https://arxiv.org/abs/2302.01318} {Accelerating large language model decoding with speculative sampling}.
\newblock \emph{arXiv preprint arXiv:2302.01318}.

\bibitem[{Devlin et~al.(2019)Devlin, Chang, Lee, and Toutanova}]{bert}
Jacob Devlin, Ming-Wei Chang, Kenton Lee, and Kristina Toutanova. 2019.
\newblock \href {https://doi.org/10.18653/v1/N19-1423} {{BERT}: Pre-training of deep bidirectional transformers for language understanding}.
\newblock In \emph{Proceedings of the 2019 Conference of the North {A}merican Chapter of the Association for Computational Linguistics: Human Language Technologies, Volume 1 (Long and Short Papers)}, pages 4171--4186, Minneapolis, Minnesota. Association for Computational Linguistics.

\bibitem[{Dubey et~al.(2024)Dubey, Jauhri, Pandey, Kadian, Al-Dahle, Letman, Mathur, Schelten, Yang, Fan et~al.}]{llama-3}
Abhimanyu Dubey, Abhinav Jauhri, Abhinav Pandey, Abhishek Kadian, Ahmad Al-Dahle, Aiesha Letman, Akhil Mathur, Alan Schelten, Amy Yang, Angela Fan, et~al. 2024.
\newblock \href {https://arxiv.org/abs/2407.21783} {The llama 3 herd of models}.
\newblock \emph{arXiv preprint arXiv:2407.21783}.

\bibitem[{Guo et~al.(2017)Guo, Pleiss, Sun, and Weinberger}]{guo2017calibration}
Chuan Guo, Geoff Pleiss, Yu~Sun, and Kilian~Q Weinberger. 2017.
\newblock \href {https://proceedings.mlr.press/v70/guo17a} {On calibration of modern neural networks}.
\newblock \emph{International Conference on Machine Learning (ICML)}.

\bibitem[{Guo et~al.(2021)Guo, Sablayrolles, J{\'e}gou, and Kiela}]{gbda}
Chuan Guo, Alexandre Sablayrolles, Herv{\'e} J{\'e}gou, and Douwe Kiela. 2021.
\newblock \href {https://doi.org/10.18653/v1/2021.emnlp-main.464} {Gradient-based adversarial attacks against text transformers}.
\newblock In \emph{Proceedings of the 2021 Conference on Empirical Methods in Natural Language Processing}, pages 5747--5757, Online and Punta Cana, Dominican Republic. Association for Computational Linguistics.

\bibitem[{Hada et~al.(2021)Hada, Sudhir, Mishra, Yannakoudakis, Mohammad, and Shutova}]{offensive-reddit}
Rishav Hada, Sohi Sudhir, Pushkar Mishra, Helen Yannakoudakis, Saif~M. Mohammad, and Ekaterina Shutova. 2021.
\newblock \href {https://doi.org/10.18653/v1/2021.acl-long.210} {Ruddit: {N}orms of offensiveness for {E}nglish {R}eddit comments}.
\newblock In \emph{Proceedings of the 59th Annual Meeting of the Association for Computational Linguistics and the 11th International Joint Conference on Natural Language Processing (Volume 1: Long Papers)}, pages 2700--2717, Online. Association for Computational Linguistics.

\bibitem[{Han et~al.(2024)Han, Rao, Ettinger, Jiang, Lin, Lambert, Choi, and Dziri}]{wildguard}
Seungju Han, Kavel Rao, Allyson Ettinger, Liwei Jiang, Bill~Yuchen Lin, Nathan Lambert, Yejin Choi, and Nouha Dziri. 2024.
\newblock \href {https://openreview.net/forum?id=Ich4tv4202#discussion} {{WildGuard}: Open one-stop moderation tools for safety risks, jailbreaks, and refusals of {LLM}s}.
\newblock \emph{Advances in Neural Information Processing Systems (NeurIPS) Datasets and Benchmarks Track}.

\bibitem[{Kingma and Ba(2015)}]{kingma2014adam}
Diederik~P Kingma and Jimmy Ba. 2015.
\newblock \href {https://arxiv.org/abs/1412.6980} {Adam: A method for stochastic optimization}.
\newblock \emph{International Conference on Learning Representations (ICLR)}.

\bibitem[{Lee(2016)}]{tay-chatbot}
Peter Lee. 2016.
\newblock \href {https://blogs.microsoft.com/blog/2016/03/25/learning-tays-introduction/} {Learning from {Tay's} introduction}.

\bibitem[{Lee et~al.(2025)Lee, Seong, Lee, Kang, Chen, Wagner, Bengio, Lee, and Hwang}]{harmaug}
Seanie Lee, Haebin Seong, Dong~Bok Lee, Minki Kang, Xiaoyin Chen, Dominik Wagner, Yoshua Bengio, Juho Lee, and Sung~Ju Hwang. 2025.
\newblock \href {https://openreview.net/forum?id=y3zswp3gek} {{HarmAug}: Effective data augmentation for knowledge distillation of safety guard models}.
\newblock \emph{International Conference on Learning Representations (ICLR)}.

\bibitem[{Leviathan et~al.(2023)Leviathan, Kalman, and Matias}]{speculative-decoding}
Yaniv Leviathan, Matan Kalman, and Yossi Matias. 2023.
\newblock \href {https://proceedings.mlr.press/v202/leviathan23a} {Fast inference from transformers via speculative decoding}.
\newblock In \emph{International Conference on Machine Learning (ICML)}.

\bibitem[{Lin et~al.(2023)Lin, Wang, Tong, Wang, Guo, Wang, and Shang}]{toxic-chat}
Zi~Lin, Zihan Wang, Yongqi Tong, Yangkun Wang, Yuxin Guo, Yujia Wang, and Jingbo Shang. 2023.
\newblock \href {https://doi.org/10.18653/v1/2023.findings-emnlp.311} {{T}oxic{C}hat: Unveiling hidden challenges of toxicity detection in real-world user-{AI} conversation}.
\newblock In \emph{Findings of the Association for Computational Linguistics: EMNLP 2023}, pages 4694--4702, Singapore. Association for Computational Linguistics.

\bibitem[{Liu et~al.(2025)Liu, Huang, Wang, Gu, and Wang}]{liu2024calibration}
Hongfu Liu, Hengguan Huang, Hao Wang, Xiangming Gu, and Ye~Wang. 2025.
\newblock \href {https://openreview.net/forum?id=wUbum0nd9N} {On calibration of {LLM}-based guard models for reliable content moderation}.
\newblock \emph{International Conference on Learning Representations (ICLR)}.

\bibitem[{Liu et~al.(2024)Liu, Xu, Chen, and Xiao}]{autodan}
Xiaogeng Liu, Nan Xu, Muhao Chen, and Chaowei Xiao. 2024.
\newblock \href {https://openreview.net/forum?id=7Jwpw4qKkb} {Auto{DAN}: Generating stealthy jailbreak prompts on aligned large language models}.
\newblock \emph{International Conference on Learning Representations (ICLR)}.

\bibitem[{Llama~Team(2024)}]{metallamaguard3}
AI~@~Meta Llama~Team. 2024.
\newblock The llama 3 family of models.
\newblock \url{https://github.com/meta-llama/PurpleLlama/blob/main/Llama-Guard3/1B/MODEL_CARD.md}.

\bibitem[{Markov et~al.(2023)Markov, Zhang, Agarwal, Nekoul, Lee, Adler, Jiang, and Weng}]{oai-ds}
Todor Markov, Chong Zhang, Sandhini Agarwal, Florentine~Eloundou Nekoul, Theodore Lee, Steven Adler, Angela Jiang, and Lilian Weng. 2023.
\newblock \href {https://dl.acm.org/doi/10.1609/aaai.v37i12.26752} {A holistic approach to undesired content detection in the real world}.
\newblock \emph{Association for the Advancement of Artificial Intelligence (AAAI)}.

\bibitem[{Mazeika et~al.(2024)Mazeika, Phan, Yin, Zou, Wang, Mu, Sakhaee, Li, Basart, Li et~al.}]{harmbench}
Mantas Mazeika, Long Phan, Xuwang Yin, Andy Zou, Zifan Wang, Norman Mu, Elham Sakhaee, Nathaniel Li, Steven Basart, Bo~Li, et~al. 2024.
\newblock \href {https://proceedings.mlr.press/v235/mazeika24a} {Harmbench: A standardized evaluation framework for automated red teaming and robust refusal}.
\newblock \emph{Internation Conference on Machine Learning (ICML)}.

\bibitem[{Mehrotra et~al.(2024)Mehrotra, Zampetakis, Kassianik, Nelson, Anderson, Singer, and Karbasi}]{tap}
Anay Mehrotra, Manolis Zampetakis, Paul Kassianik, Blaine Nelson, Hyrum Anderson, Yaron Singer, and Amin Karbasi. 2024.
\newblock \href {https://openreview.net/forum?id=SoM3vngOH5} {Tree of attacks: Jailbreaking black-box {LLM}s automatically}.
\newblock \emph{Advances in Neural Information Processing systems (NeurIPS)}.

\bibitem[{Nair and Hinton(2010)}]{Nair2010RectifiedLU}
Vinod Nair and Geoffrey~E. Hinton. 2010.
\newblock \href {https://icml.cc/Conferences/2010/papers/432.pdf} {Rectified linear units improve restricted boltzmann machines}.
\newblock In \emph{International Conference on Machine Learning (ICML)}.

\bibitem[{Ouyang et~al.(2022)Ouyang, Wu, Jiang, Almeida, Wainwright, Mishkin, Zhang, Agarwal, Slama, Ray et~al.}]{rlhf}
Long Ouyang, Jeffrey Wu, Xu~Jiang, Diogo Almeida, Carroll Wainwright, Pamela Mishkin, Chong Zhang, Sandhini Agarwal, Katarina Slama, Alex Ray, et~al. 2022.
\newblock \href {https://openreview.net/forum?id=TG8KACxEON} {Training language models to follow instructions with human feedback}.
\newblock \emph{Advances in Neural Information Processing systems (NeurIPS)}.

\bibitem[{Padhi et~al.(2024)Padhi, Nagireddy, Cornacchia, Chaudhury, Pedapati, Dognin, Murugesan, Miehling, Cooper, Fraser, Zizzo, Hameed, Purcell, Desmond, Pan, Ashktorab, Vejsbjerg, Daly, Hind, Geyer, Rawat, Varshney, and Sattigeri}]{granite-guardian}
Inkit Padhi, Manish Nagireddy, Giandomenico Cornacchia, Subhajit Chaudhury, Tejaswini Pedapati, Pierre Dognin, Keerthiram Murugesan, Erik Miehling, Martín~Santillán Cooper, Kieran Fraser, Giulio Zizzo, Muhammad~Zaid Hameed, Mark Purcell, Michael Desmond, Qian Pan, Zahra Ashktorab, Inge Vejsbjerg, Elizabeth~M. Daly, Michael Hind, Werner Geyer, Ambrish Rawat, Kush~R. Varshney, and Prasanna Sattigeri. 2024.
\newblock Granite guardian.
\newblock \emph{arXiv preprint arXiv:2412.07724}.

\bibitem[{Paszke et~al.(2019)Paszke, Gross, Massa, Lerer, Bradbury, Chanan, Killeen, Lin, Gimelshein, Antiga et~al.}]{pytorch}
Adam Paszke, Sam Gross, Francisco Massa, Adam Lerer, James Bradbury, Gregory Chanan, Trevor Killeen, Zeming Lin, Natalia Gimelshein, Luca Antiga, et~al. 2019.
\newblock \href {https://papers.nips.cc/paper_files/paper/2019/hash/bdbca288fee7f92f2bfa9f7012727740-Abstract.html} {Pytorch: An imperative style, high-performance deep learning library}.
\newblock \emph{Advances in neural information processing systems (NeurIPS)}.

\bibitem[{R{\"o}ttger et~al.(2024)R{\"o}ttger, Kirk, Vidgen, Attanasio, Bianchi, and Hovy}]{xstest}
Paul R{\"o}ttger, Hannah Kirk, Bertie Vidgen, Giuseppe Attanasio, Federico Bianchi, and Dirk Hovy. 2024.
\newblock \href {https://doi.org/10.18653/v1/2024.naacl-long.301} {{XST}est: A test suite for identifying exaggerated safety behaviours in large language models}.
\newblock In \emph{Proceedings of the 2024 Conference of the North American Chapter of the Association for Computational Linguistics: Human Language Technologies (Volume 1: Long Papers)}, pages 5377--5400, Mexico City, Mexico. Association for Computational Linguistics.

\bibitem[{Sethupathy(2024)}]{llm-risk}
Guru Sethupathy. 2024.
\newblock \href {https://fairnow.ai/executives-guide-risks-of-llms/} {An executive’s guide to the risks of large language models ({LLMs}): From hallucinations to copyright infringement}.

\bibitem[{Shin et~al.(2020)Shin, Razeghi, Logan~IV, Wallace, and Singh}]{autoprompt}
Taylor Shin, Yasaman Razeghi, Robert~L. Logan~IV, Eric Wallace, and Sameer Singh. 2020.
\newblock \href {https://doi.org/10.18653/v1/2020.emnlp-main.346} {{A}uto{P}rompt: {E}liciting {K}nowledge from {L}anguage {M}odels with {A}utomatically {G}enerated {P}rompts}.
\newblock In \emph{Proceedings of the 2020 Conference on Empirical Methods in Natural Language Processing (EMNLP)}, pages 4222--4235, Online. Association for Computational Linguistics.

\bibitem[{Su et~al.(2024)Su, Ahmed, Lu, Pan, Bo, and Liu}]{rope}
Jianlin Su, Murtadha Ahmed, Yu~Lu, Shengfeng Pan, Wen Bo, and Yunfeng Liu. 2024.
\newblock \href {https://www.sciencedirect.com/science/article/abs/pii/S0925231223011864} {Roformer: Enhanced transformer with rotary position embedding}.
\newblock \emph{Neurocomputing}, 568:127063.

\bibitem[{Vidgen et~al.(2021)Vidgen, Thrush, Waseem, and Kiela}]{roberta-hate-speech}
Bertie Vidgen, Tristan Thrush, Zeerak Waseem, and Douwe Kiela. 2021.
\newblock \href {https://doi.org/10.18653/v1/2021.acl-long.132} {Learning from the worst: Dynamically generated datasets to improve online hate detection}.
\newblock In \emph{Proceedings of the 59th Annual Meeting of the Association for Computational Linguistics and the 11th International Joint Conference on Natural Language Processing (Volume 1: Long Papers)}, pages 1667--1682, Online. Association for Computational Linguistics.

\bibitem[{Wagner et~al.(2024)Wagner, Lee, Baumann, Seeberger, Riedhammer, and Bocklet}]{speculative-decoding-3}
Dominik Wagner, Seanie Lee, Ilja Baumann, Philipp Seeberger, Korbinian Riedhammer, and Tobias Bocklet. 2024.
\newblock \href {https://doi.org/10.18653/v1/2024.emnlp-main.370} {Optimized speculative sampling for {GPU} hardware accelerators}.
\newblock In \emph{Proceedings of the 2024 Conference on Empirical Methods in Natural Language Processing}, pages 6442--6458, Miami, Florida, USA. Association for Computational Linguistics.

\bibitem[{Wallace et~al.(2019)Wallace, Feng, Kandpal, Gardner, and Singh}]{uat}
Eric Wallace, Shi Feng, Nikhil Kandpal, Matt Gardner, and Sameer Singh. 2019.
\newblock \href {https://doi.org/10.18653/v1/D19-1221} {Universal adversarial triggers for attacking and analyzing {NLP}}.
\newblock In \emph{Proceedings of the 2019 Conference on Empirical Methods in Natural Language Processing and the 9th International Joint Conference on Natural Language Processing (EMNLP-IJCNLP)}, pages 2153--2162, Hong Kong, China. Association for Computational Linguistics.

\bibitem[{Warner et~al.(2024)Warner, Chaffin, Clavi{\'e}, Weller, Hallstr{\"o}m, Taghadouini, Gallagher, Biswas, Ladhak, Aarsen et~al.}]{modernbert}
Benjamin Warner, Antoine Chaffin, Benjamin Clavi{\'e}, Orion Weller, Oskar Hallstr{\"o}m, Said Taghadouini, Alexis Gallagher, Raja Biswas, Faisal Ladhak, Tom Aarsen, et~al. 2024.
\newblock \href {https://arxiv.org/abs/2412.13663} {Smarter, better, faster, longer: A modern bidirectional encoder for fast, memory efficient, and long context finetuning and inference}.
\newblock \emph{arXiv preprint arXiv:2412.13663}.

\bibitem[{Wolf et~al.(2020)Wolf, Debut, Sanh, Chaumond, Delangue, Moi, Cistac, Rault, Louf, Funtowicz, Davison, Shleifer, von Platen, Ma, Jernite, Plu, Xu, Le~Scao, Gugger, Drame, Lhoest, and Rush}]{transformers}
Thomas Wolf, Lysandre Debut, Victor Sanh, Julien Chaumond, Clement Delangue, Anthony Moi, Pierric Cistac, Tim Rault, Remi Louf, Morgan Funtowicz, Joe Davison, Sam Shleifer, Patrick von Platen, Clara Ma, Yacine Jernite, Julien Plu, Canwen Xu, Teven Le~Scao, Sylvain Gugger, Mariama Drame, Quentin Lhoest, and Alexander Rush. 2020.
\newblock \href {https://doi.org/10.18653/v1/2020.emnlp-demos.6} {Transformers: State-of-the-art natural language processing}.
\newblock In \emph{Proceedings of the 2020 Conference on Empirical Methods in Natural Language Processing: System Demonstrations}, pages 38--45, Online. Association for Computational Linguistics.

\bibitem[{Yuan et~al.(2024)Yuan, Jiao, Wang, tse Huang, He, Shi, and Tu}]{yuan2024cipherchat}
Youliang Yuan, Wenxiang Jiao, Wenxuan Wang, Jen tse Huang, Pinjia He, Shuming Shi, and Zhaopeng Tu. 2024.
\newblock \href {https://openreview.net/forum?id=MbfAK4s61A} {{GPT}-4 is too smart to be safe: Stealthy chat with {LLM}s via cipher}.
\newblock \emph{International Conference on Learning Representations ({ICLR})}.

\bibitem[{Zeng et~al.(2024)Zeng, Lin, Zhang, Yang, Jia, and Shi}]{pap}
Yi~Zeng, Hongpeng Lin, Jingwen Zhang, Diyi Yang, Ruoxi Jia, and Weiyan Shi. 2024.
\newblock \href {https://doi.org/10.18653/v1/2024.acl-long.773} {How johnny can persuade {LLM}s to jailbreak them: Rethinking persuasion to challenge {AI} safety by humanizing {LLM}s}.
\newblock In \emph{Proceedings of the 62nd Annual Meeting of the Association for Computational Linguistics (Volume 1: Long Papers)}, pages 14322--14350, Bangkok, Thailand. Association for Computational Linguistics.

\bibitem[{Zhao et~al.(2021)Zhao, Wallace, Feng, Klein, and Singh}]{zhao2021calibrate}
Zihao Zhao, Eric Wallace, Shi Feng, Dan Klein, and Sameer Singh. 2021.
\newblock \href {https://proceedings.mlr.press/v139/zhao21c} {Calibrate before use: Improving few-shot performance of language models}.
\newblock \emph{International Conference on Machine Learning (ICML)}.

\bibitem[{Zhou et~al.(2024)Zhou, Wan, Proleev, Mincu, Chen, Heller, and Roy}]{zhou2023batch}
Han Zhou, Xingchen Wan, Lev Proleev, Diana Mincu, Jilin Chen, Katherine~A Heller, and Subhrajit Roy. 2024.
\newblock \href {https://openreview.net/forum?id=L3FHMoKZcS} {Batch calibration: Rethinking calibration for in-context learning and prompt engineering}.
\newblock \emph{International Conference on Learning Representations (ICLR)}.

\bibitem[{Zou et~al.(2023)Zou, Wang, Kolter, and Fredrikson}]{zou2023universal}
Andy Zou, Zifan Wang, J.~Zico Kolter, and Matt Fredrikson. 2023.
\newblock \href {https://arxiv.org/abs/2307.15043} {Universal and transferable adversarial attacks on aligned language models}.
\newblock \emph{arXiv preprint arXiv: 2307.15043}.

\end{thebibliography}
\clearpage
\appendix
\label{sec:appendix}

\section{\texorpdfstring{Proof of~\Cref{thm}}{Proof Theorem 3.1}}
\label{app:proof}
\begin{proof}
\begin{align*}
    R_\text{adaptive} - R_\text{oracle} = &\mathbb{E}[I(\rvx,\rvy)\ell(p(\rvx,\rvy),c) \\
    &+ (1-I(\rvx,\rvy))\ell(q(\rvx,\rvy),c) \\
    &- t(\rvx,\rvy)\ell(p(\rvx,\rvy),c) \\
    & - (1-t(\rvx,\rvy))\ell(q(\rvx,\rvy),c)].
\end{align*}
Taking the absolute value and using the fact that
\begin{equation*}
    \lvert I(\rvx,\rvy) -t(\rvx,\rvy)\rvert = \one_{\{ I(\rvx,\rvy) \neq t(\rvx,\rvy)\}},
\end{equation*}
we obtain the following inequality,
\begin{align*}
    &\lvert R_\text{adaptive} - R_\text{oracle} \rvert \\
    &\leq \mathbb{E}[
    \one_{\{I(\rvx,\rvy) \neq t(\rvx,\rvy) \}} \lvert \ell(p(\rvx,\rvy),c) - \ell(q(\rvx,\rvy),c)\rvert
    ].
\end{align*}
For notational brevity, we use $I\neq t$ to denote $I(\rvx,\rvy)\neq t(\rvx,\rvy)$. By applying the Cauchy-Schwarz inequality, we obtain the final result,
\begin{align*}
    &\lvert R_\text{adaptive} - R_\text{oracle} \rvert \\&\leq \sqrt{\mathbb{E}[
    \one^2_{\{I \neq t \}}]} \sqrt{\mathbb{E}[ \lvert \ell(p(\rvx,\rvy),c) - \ell(q(\rvx,\rvy),c)\rvert^2 
    ]} \\
    &= \sqrt{\mathbb{E}[\one_{\{I(\rvx,\rvy) \neq t(\rvx,\rvy)\}}]} M\\
    &= \sqrt{\mathbb{P}(I(\rvx,\rvy)\neq t(\rvx,\rvy))} M
\end{align*}
where $M=\sqrt{\mathbb{E}[ \lvert \ell(p(\rvx,\rvy),c) - \ell(q(\rvx,\rvy),c)\rvert^2 
    ]}$. Thus, we have
\begin{equation*}
    R_\text{adaptive} \leq R_\text{oracle} + M \sqrt{\mathbb{P}(I(\rvx,\rvy) \neq t(\rvx,\rvy))}.
\end{equation*}

\end{proof}

\section{Data Statistics}
\label{sec:data_statistics}

\begin{table}[ht]
    \centering
    \resizebox{1.0\columnwidth}{!}{\begin{tabular}{lccc}
    \toprule
    \textbf{Dataset}     & \# of safe & \# of harmful & Total \\
    \midrule
    OAI     & 1,158 & \phantom{0}522  & 1,680\\
    WildGuardMix & 1,407 &  \phantom{0}282& 1,689\\
    WildGuardMix-p & \phantom{0,}945 & \phantom{0}754 & 1,699 \\
    ToxicChat & 4,721 & \phantom{0}362 & 5,083\\
    XSTest & \phantom{0,}368 & \phantom{0}\phantom{0}78 & \phantom{0,}446 \\
    Harmbench & \phantom{0,}329 & \phantom{0}273 & \phantom{0,}602 \\
    \bottomrule
    \end{tabular}}
    \caption{Statistics of each dataset.}
    \label{tab:data-stat}
\end{table}

\section{Safety Guard Models}
\label{sec:model}
We use PyTorch~\citep{pytorch} and Transformers~\citep{transformers} to implement all methods. All the pre-trained models, including safety guard models, used for our experiments are available in Hugging Face Hub. We list the identifier and link for each model on the Hugging Face Hub in~\Cref{tab:model}.

\begin{table}[t]
    \centering
    \resizebox{0.98\columnwidth}{!}{\begin{tabular}{lc}
    \toprule
    \textbf{Model} & \textbf{Hugging Face Hub Identifier} \\
    \midrule
    \rowcolor{rowgray}
     \texttt{Llama-Guard-3-1B}    & \href{https://huggingface.co/meta-llama/Llama-Guard-3-1B}{meta-llama/Llama-Guard-3-1B}  \\
    \texttt{Llama-Guard-3-8B}     & \href{https://huggingface.co/meta-llama/Llama-Guard-3-8B}{meta-llama/Llama-Guard-3-8B}  \\
    \rowcolor{rowgray}
    \texttt{Granite-Guardian-3-8B} & \href{https://huggingface.co/ibm-granite/granite-guardian-3.0-8b}{ibm-granite/granite-guardian-3.0-8b}\\
    \texttt{ModernBert} & \href{https://huggingface.co/answerdotai/ModernBERT-large}{answerdotai/ModernBERT-large} \\
    \rowcolor{rowgray}
    \texttt{Llama-3.1-8B-Instruct} & \href{https://huggingface.co/meta-llama/Llama-3.1-8B-Instruct}{meta-llama/Llama-3.1-8B-Instruct}\\
    \bottomrule
    \end{tabular}
    }
    \caption{Hugging Face Hub model identifiers for the pre-trained models used in our work.}
    \label{tab:model}
\end{table}

\begin{sidewaystable}[ht]
\centering
\scriptsize
\setlength{\tabcolsep}{3pt}
\caption{The average of safety F1 score, precision (Prec.), recall (Rec.), and latency (Lat.) when using smaller (\texttt{Llama-Guard-3-1B}) and larger (\texttt{Llama-Guard-3-8B}) models.}
\label{tab:full_8B}
\resizebox{\textwidth}{!}{%
\begin{tabular}{l|cccc|cccc|cccc|cccc|cccc|cccc}
\toprule
\textbf{Method} & \multicolumn{4}{c|}{\textbf{WildGuardMix-p}} & \multicolumn{4}{c|}{\textbf{ToxicChat}} & \multicolumn{4}{c|}{\textbf{OAI}} & \multicolumn{4}{c|}{\textbf{WildGuardMix}} & \multicolumn{4}{c|}{\textbf{XSTest}} & \multicolumn{4}{c}{\textbf{HarmBench}} \\
& F1 & Prec. & Rec. & Lat. & F1 & Prec. & Rec. & Lat. & F1 & Prec. & Rec. & Lat. & F1 & Prec. & Rec. & Lat. & F1 & Prec. & Rec. & Lat. & F1 & Prec. & Rec. & Lat. \\
\midrule
\rowcolor{rowgray}Small & 0.752 & 0.827 & 0.690 & 6.66 & 0.341 & 0.242 & 0.572 & 17.05 & 0.679 & 0.587 & 0.805 & 9.24 & 0.664 & 0.677 & 0.653 & 14.60 & 0.853 & 0.889 & 0.821 & 3.59 & 0.785 & 0.726 & 0.854 & 5.45 \\
Large & 0.771 & 0.929 & 0.659 & 22.67 & 0.487 & 0.467 & 0.508 & 70.68 & 0.787 & 0.784 & 0.791 & 42.72 & 0.705 & 0.811 & 0.624 & 64.16 & 0.904 & 0.971 & 0.846 & 10.54 & 0.846 & 0.811 & 0.883 & 25.63 \\
\rowcolor{rowgray}Random & 0.758 & 0.882 & 0.664 & 17.79 & 0.399 & 0.320 & 0.534 & 59.23 & 0.732 & 0.679 & 0.796 & 28.42 & 0.684 & 0.753 & 0.629 & 42.18 & 0.883 & 0.932 & 0.839 & 8.38 & 0.813 & 0.770 & 0.863 & 16.17 \\
Ent & 0.764 & 0.902 & 0.662 & 16.16 & 0.473 & 0.421 & 0.539 & 41.49 & 0.763 & 0.736 & 0.791 & 22.35 & 0.715 & 0.824 & 0.631 & 31.64 & 0.892 & 0.943 & 0.846 & 6.14 & 0.846 & 0.809 & 0.886 & 13.09 \\
\rowcolor{rowgray}TS & 0.771 & 0.932 & 0.658 & 25.39 & 0.485 & 0.470 & 0.500 & 76.68 & 0.791 & 0.795 & 0.787 & 48.36 & 0.707 & 0.822 & 0.621 & 73.26 & 0.904 & 0.971 & 0.846 & 11.07 & 0.847 & 0.816 & 0.879 & 30.17 \\
CC & 0.776 & 0.892 & 0.687 & 14.75 & 0.461 & 0.387 & 0.569 & 42.54 & 0.760 & 0.707 & 0.820 & 22.53 & 0.730 & 0.807 & 0.667 & 33.83 & 0.892 & 0.943 & 0.846 & 5.80 & 0.838 & 0.791 & 0.890 & 12.49 \\
\rowcolor{rowgray}BC & 0.782 & 0.868 & 0.712 & 14.97 & 0.426 & 0.328 & 0.605 & 45.01 & 0.734 & 0.647 & 0.847 & 32.23 & 0.726 & 0.739 & 0.713 & 39.37 & 0.882 & 0.905 & 0.859 & 6.02 & 0.836 & 0.769 & 0.916 & 11.78 \\
SafeRoute (Ours) & 0.782 & 0.926 & 0.677 & 11.24 & 0.483 & 0.416 & 0.576 & 31.51 & 0.750 & 0.672 & 0.848 & 18.46 & 0.734 & 0.856 & 0.643 & 28.18 & 0.897 & 0.967 & 0.836 & 4.71 & 0.843 & 0.797 & 0.895 & 9.92 \\
\rowcolor{rowgray}Oracle & 0.837 & 0.951 & 0.748 & 28.27 & 0.609 & 0.582 & 0.638 & 95.18 & 0.859 & 0.842 & 0.877 & 63.58 & 0.810 & 0.904 & 0.734 & 77.19 & 0.932 & 0.986 & 0.885 & 12.47 & 0.885 & 0.834 & 0.941 & 31.25 \\
\bottomrule
\end{tabular}
}
\end{sidewaystable}

\begin{sidewaystable}[ht]
\centering
\scriptsize
\setlength{\tabcolsep}{3pt}
\caption{The average of safety F1 score, precision (Prec.), recall (Rec.), and latency (Lat.) when using smaller (\texttt{Llama-Guard-3-1B}) and larger (\texttt{Granite-Guardian-3-8B}) models.}
\label{tab:full_guardian}
\resizebox{\textwidth}{!}{%
\begin{tabular}{l|cccc|cccc|cccc|cccc|cccc|cccc}
\toprule
\textbf{Method} & \multicolumn{4}{c|}{\textbf{WildGuardMix-p}} & \multicolumn{4}{c|}{\textbf{ToxicChat}} & \multicolumn{4}{c|}{\textbf{OAI}} & \multicolumn{4}{c|}{\textbf{WildGuardMix}} & \multicolumn{4}{c|}{\textbf{XSTest}} & \multicolumn{4}{c}{\textbf{HarmBench}} \\
& F1 & Prec. & Rec. & Lat. & F1 & Prec. & Rec. & Lat. & F1 & Prec. & Rec. & Lat. & F1 & Prec. & Rec. & Lat. & F1 & Prec. & Rec. & Lat. & F1 & Prec. & Rec. & Lat. \\
\midrule
\rowcolor{rowgray}Small & 0.752 & 0.827 & 0.690 & 6.46 & 0.341 & 0.242 & 0.570 & 17.20 & 0.679 & 0.587 & 0.805 & 9.32 & 0.664 & 0.677 & 0.653 & 14.90 & 0.853 & 0.889 & 0.821 & 3.47 & 0.785 & 0.726 & 0.854 & 5.60 \\
Large & 0.832 & 0.757 & 0.923 & 25.65 & 0.567 & 0.423 & 0.859 & 93.55 & 0.731 & 0.617 & 0.897 & 62.64 & 0.751 & 0.770 & 0.734 & 87.94 & 0.857 & 0.913 & 0.808 & 12.85 & 0.820 & 0.848 & 0.795 & 34.47 \\
\rowcolor{rowgray}Random & 0.795 & 0.795 & 0.796 & 21.41 & 0.461 & 0.338 & 0.716 & 52.70 & 0.712 & 0.618 & 0.847 & 33.87 & 0.712 & 0.743 & 0.682 & 53.06 & 0.861 & 0.893 & 0.813 & 9.36 & 0.797 & 0.782 & 0.812 & 19.69 \\
Ent & 0.824 & 0.837 & 0.812 & 19.61 & 0.528 & 0.418 & 0.716 & 43.52 & 0.743 & 0.661 & 0.849 & 25.93 & 0.751 & 0.795 & 0.713 & 37.93 & 0.889 & 0.970 & 0.821 & 6.12 & 0.844 & 0.835 & 0.854 & 15.31 \\
\rowcolor{rowgray}TS & 0.843 & 0.780 & 0.916 & 28.64 & 0.576 & 0.440 & 0.832 & 83.49 & 0.743 & 0.641 & 0.885 & 62.34 & 0.749 & 0.788 & 0.713 & 97.48 & 0.855 & 0.925 & 0.795 & 12.90 & 0.805 & 0.841 & 0.773 & 38.77 \\
CC & 0.830 & 0.817 & 0.842 & 16.57 & 0.506 & 0.381 & 0.751 & 45.42 & 0.730 & 0.629 & 0.872 & 25.76 & 0.747 & 0.761 & 0.734 & 40.84 & 0.889 & 0.970 & 0.821 & 5.94 & 0.843 & 0.806 & 0.883 & 14.74 \\
\rowcolor{rowgray}BC & 0.833 & 0.798 & 0.871 & 16.77 & 0.468 & 0.333 & 0.790 & 47.12 & 0.707 & 0.584 & 0.895 & 25.51 & 0.742 & 0.716 & 0.769 & 46.90 & 0.874 & 0.904 & 0.846 & 6.47 & 0.833 & 0.766 & 0.912 & 13.65 \\
SafeRoute (Ours) & 0.848 & 0.879 & 0.819 & 12.19 & 0.525 & 0.395 & 0.781 & 35.12 & 0.715 & 0.589 & 0.910 & 21.02 & 0.753 & 0.825 & 0.693 & 25.07 & 0.892 & 0.970 & 0.826 & 4.28 & 0.834 & 0.796 & 0.876 & 10.32 \\
\rowcolor{rowgray}Oracle & 0.932 & 0.924 & 0.939 & 30.81 & 0.748 & 0.650 & 0.881 & 94.17 & 0.865 & 0.802 & 0.939 & 85.14 & 0.855 & 0.912 & 0.805 & 104.22 & 0.919 & 0.971 & 0.872 & 16.68 & 0.905 & 0.885 & 0.927 & 40.13 \\
\bottomrule
\end{tabular}
}
\end{sidewaystable}

\begin{figure*}[t]
\centering
\includegraphics[width=0.95\textwidth]{images/legend.pdf}
\medskip
\vspace{-0.15in}
\includegraphics[width=0.99\textwidth]{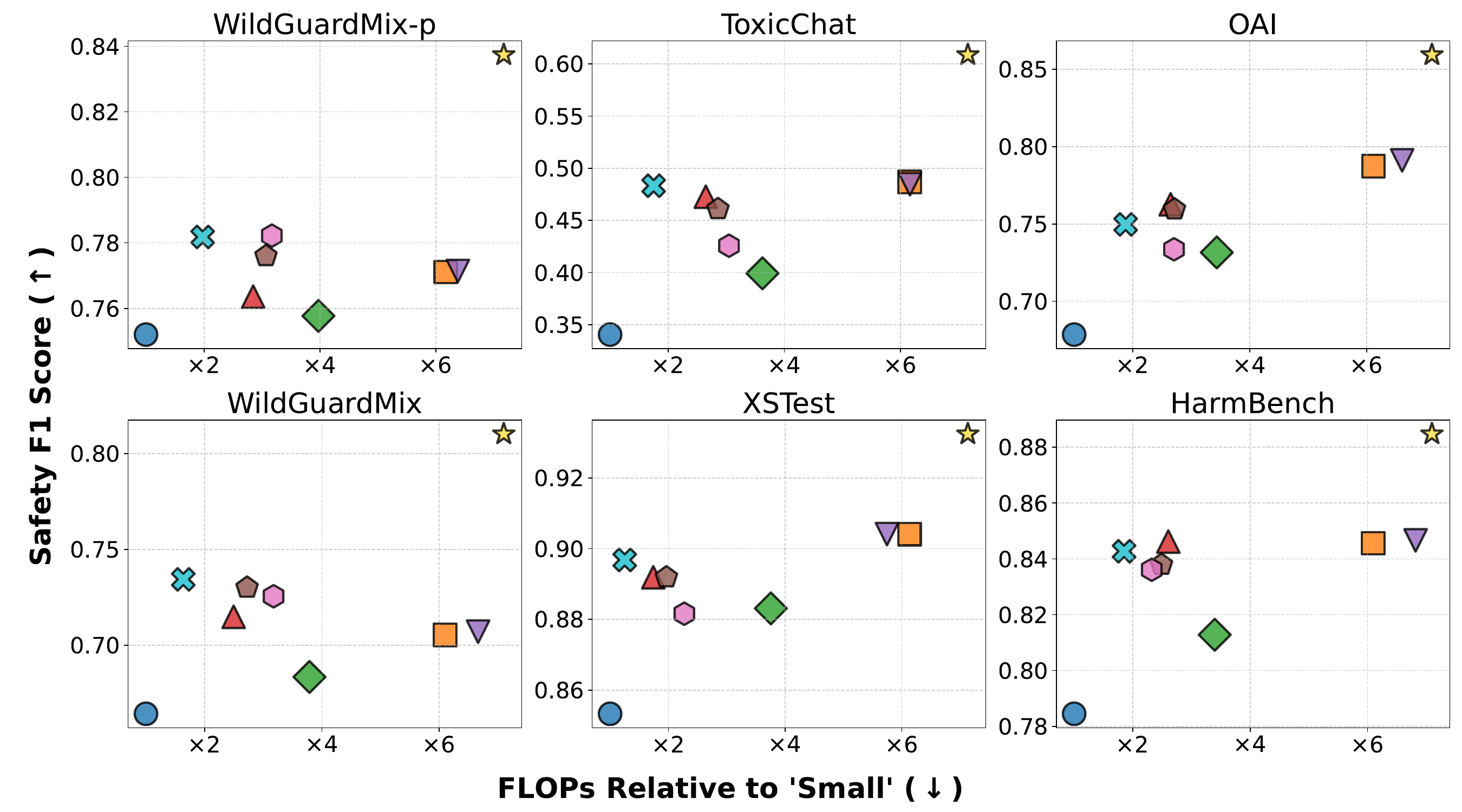}
\vspace{-0.1em}
\caption{\textbf{FLOPs ($\downarrow$) vs. safety F1 score ($\uparrow$) trade-off} when using the smaller (\texttt{Llama-Guard-3-1B}) and larger (\texttt{Llama-Guard-3-8B}) models.}
\label{fig:flops}
\end{figure*}
\begin{figure*}[t]
\centering
\includegraphics[width=0.95\textwidth]{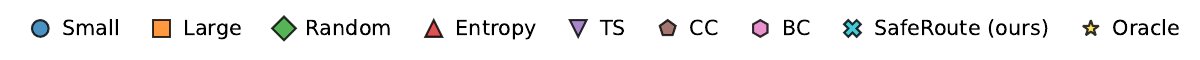}
\medskip
\vspace{-0.15in}
\includegraphics[width=0.99\textwidth]{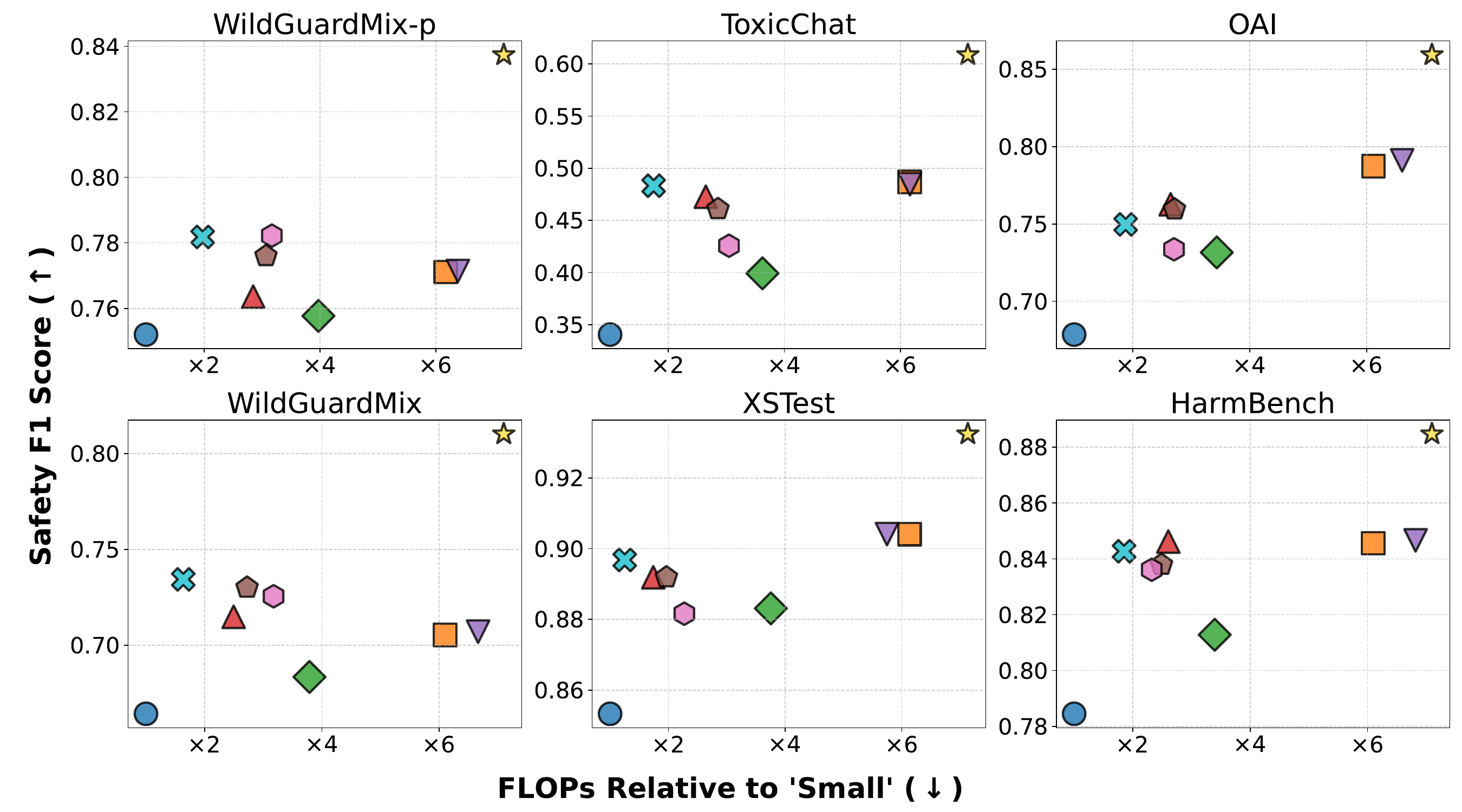}
\vspace{-0.1em}
\caption{\textbf{FLOPs ($\downarrow$) vs. safety F1 score ($\uparrow$) trade-off} when using the smaller (\texttt{Llama-Guard-3-1B}) and larger (\texttt{Granite-Guardian-3-8B}) models.}
\label{fig:flops_guardian}
\end{figure*}
\begin{figure*}[t]
\centering
\includegraphics[width=0.95\textwidth]{images/legend.pdf}
\medskip
\vspace{-0.15in}
\includegraphics[width=0.99\textwidth]{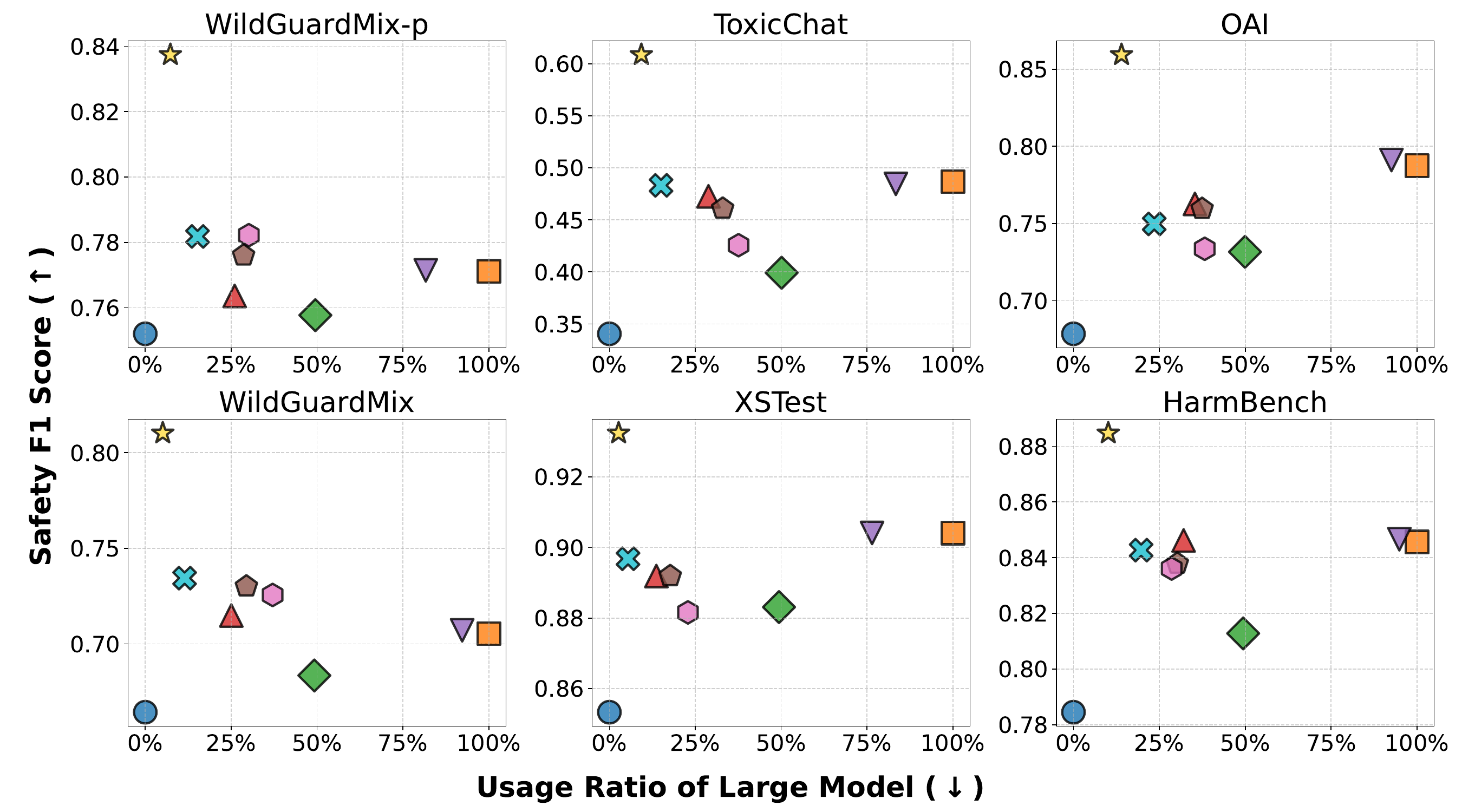}
\vspace{-0.1em}
\caption{\textbf{Usage ratio of large model ($\downarrow$) vs. safety F1 score ($\uparrow$) trade-off} when using the smaller (\texttt{Llama-Guard-3-1B}) and larger (\texttt{Llama-Guard-3-8B}) models.}
\label{fig:large_ratio}
\end{figure*}
\begin{figure*}[t]
\centering
\includegraphics[width=0.95\textwidth]{images/legend.pdf}
\medskip
\vspace{-0.15in}
\includegraphics[width=0.99\textwidth]{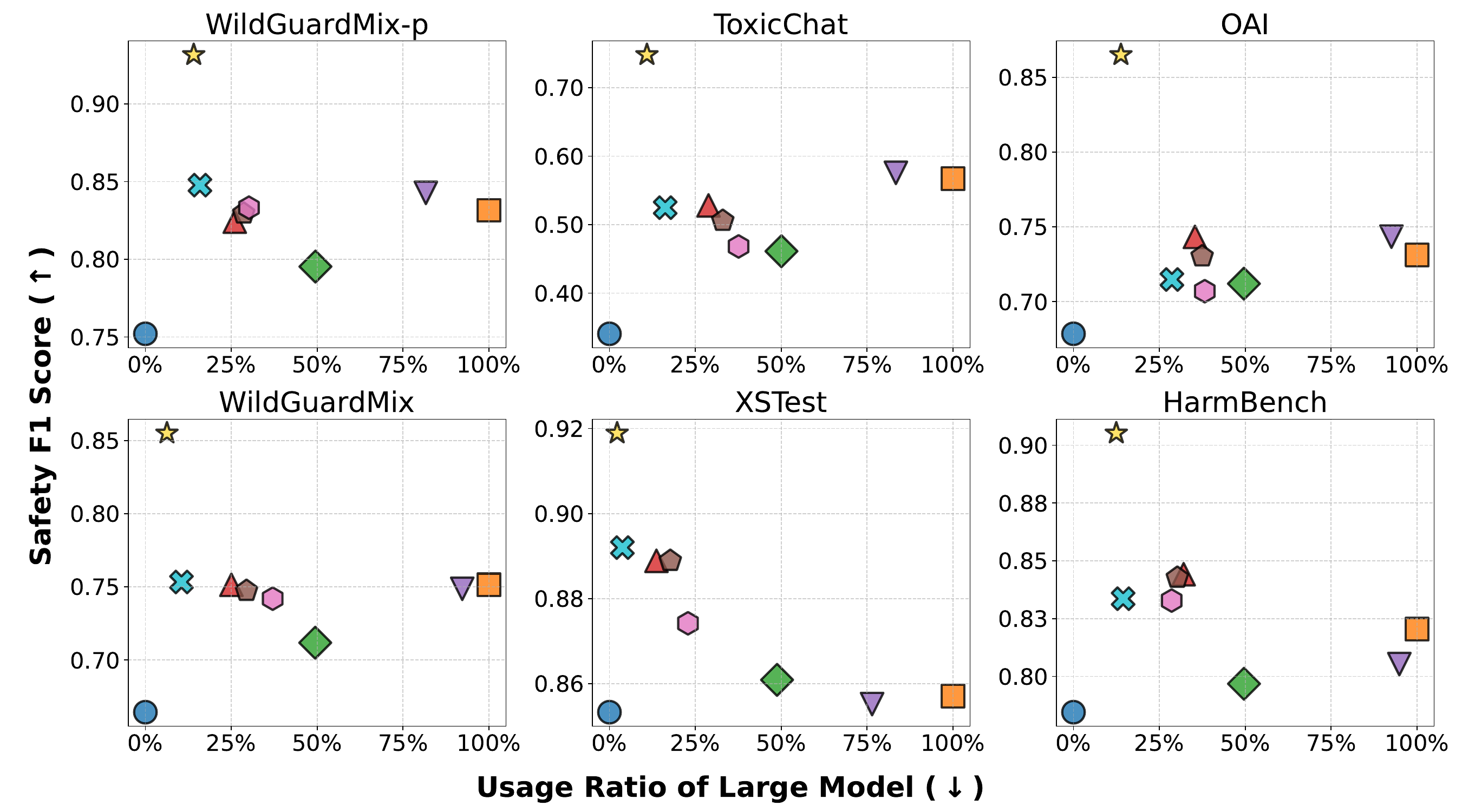}
\vspace{-0.1em}
\caption{\textbf{Usage ratio of large model ($\downarrow$) vs. safety F1 score ($\uparrow$) trade-off} using the smaller (\texttt{Llama-Guard-3-1B}) and larger (\texttt{Granite-Guardian-3-8B}) models.}
\label{fig:large_ratio_guardian}
\end{figure*}
\section{Additional Experimental Results}
\label{sec:additional_experimental_results}

In \Cref{fig:flops} and \Cref{fig:flops_guardian}, we present trade-off between FLOPs and F1 score when adaptively selecting between  the smaller (\texttt{Llama-Guard-
3-1B}) and larger (\texttt{Llama-Guard-3-8B} and \texttt{Granite-Guardian-3-8B}, respectively) models. 
In \Cref{fig:large_ratio} and \Cref{fig:large_ratio_guardian}, we present trade-off between usage ratio of large model and F1 score when adaptively selecting between  the smaller (\texttt{Llama-Guard-
3-1B}) and larger (\texttt{Llama-Guard-3-8B} and \texttt{Granite-Guardian-3-8B}, respectively) models. 

\begin{figure*}[t]
    \centering
    \resizebox{0.95\linewidth}{!}{\begin{prompt}{Prompt}
        \footnotesize
        \textbf{USER}: Paraphrase the following prompt and response respectively while preserving their original semantics. Adhere strictly to the following format. First do not include original prompt and response.  Second, start the paraphrased prompt with ``prompt:''. Third, start the paraphrased response with ``response:''. Fourth, if response is given as None, just provide it as None. \\
        prompt: \{\texttt{prompt}\} \\
        response: \{\texttt{response}\} \\
        \textbf{ASSISTANT}: \\
\end{prompt}}    
    \caption{The prompt format for paraphrasing prompt-response pairs.}
    \label{fig:prompt}
\end{figure*}

\section{Prompt for Paraphrasing}
\label{app:para}
We present the prompt format for paraphrasing prompt-response pairs in \Cref{fig:prompt}.

\end{document}